\documentclass{article}

\usepackage{PRIMEarxiv}

\usepackage[utf8]{inputenc} 
\usepackage[T1]{fontenc}    
\usepackage{hyperref}       
\usepackage{url}            
\usepackage{booktabs}       
\usepackage{amsfonts}       
\usepackage{nicefrac}       
\usepackage{microtype}      
\usepackage{fancyhdr}       
\usepackage{graphicx}       
\graphicspath{{media/}}     
\usepackage{algorithm}
\usepackage{algorithmic}

\usepackage{amsmath}
\usepackage{amssymb}
\usepackage{mathtools}
\usepackage{amsthm}
\usepackage{diagbox}
\usepackage{makecell}
\usepackage{multirow}
\usepackage{multicol}
\usepackage[table,xcdraw]{xcolor}
\usepackage{threeparttable}

\theoremstyle{definition}

\theoremstyle{remark}

\usepackage[normalem]{ulem}
\useunder{\uline}{\ul}{}
\title{RobustFair: Adversarial Evaluation through Fairness Confusion Directed Gradient Search}
\author{
   Xuran Li\textsuperscript{\rm 1,\rm 2},
   Peng Wu \textsuperscript{\rm 1,\rm 2}
   \thanks{Corresponding author: Peng Wu},
   Kaixiang Dong\textsuperscript{\rm 3,\rm 1},
   Zhen Zhang \textsuperscript{\rm 1,\rm 2}
   Yanting Chen \textsuperscript{\rm 1,\rm 2}\\ 
   \textsuperscript{\rm 1} State Key Laboratory of Computer Science, Institute of Software, Chinese Academy of Sciences, Beijing, China\\
    \textsuperscript{\rm 2} University of Chinese Academy of Sciences, Beijing, China \\
    \textsuperscript{\rm 3} Hangzhou Institute for Advanced Study, University of Chinese Academy of Sciences, Hangzhou, China \\
    \texttt{{lixr,wp,zhangzhen19,chenyt}@ios.ac.cn, dongkaixiang22@mails.ucas.ac.cn} \\
}

\begin{document}
\maketitle

\begin{abstract}
Deep neural networks (DNNs) often face challenges due to their vulnerability to various adversarial perturbations, including false perturbations that undermine prediction accuracy and biased perturbations that cause biased predictions for similar inputs. This paper introduces a novel approach called RobustFair to evaluate the accurate fairness of DNNs when subjected to these false or biased perturbations. RobustFair employs the notion of the fairness confusion matrix induced in accurate fairness to identify the crucial input features for perturbations. This matrix categorizes predictions as true fair, true biased, false fair, and false biased, and the perturbations guided by it can produce a dual impact on instances and their similar counterparts to either undermine prediction accuracy (robustness) or cause biased predictions (individual fairness). RobustFair then infers the ground truth of these generated adversarial instances based on their loss function values approximated by the total derivative. To leverage the generated instances for trustworthiness improvement, RobustFair further proposes a data augmentation strategy to prioritize adversarial instances resembling the original training set, for data augmentation and model retraining. Notably, RobustFair excels at detecting intertwined issues of robustness and individual fairness, which are frequently overlooked in standard robustness and individual fairness evaluations. This capability empowers RobustFair to enhance both robustness and individual fairness evaluations by concurrently identifying defects in either domain. Empirical case studies and quantile regression analyses on benchmark datasets demonstrate the effectiveness of the fairness confusion matrix guided perturbation for false or biased adversarial instance generation. Compared to state-of-the-art white-box robustness and individual fairness testing approaches, RobustFair identifies significantly more false adversarial instances (3.365-8.183 times) and biased adversarial instances (2.217-7.818 times), as well as numerous false or biased adversarial instances (4.807-15.885 times) previously overlooked by the other testing approaches. Furthermore, the adversarial instances identified by RobustFair outperform those identified by the other testing approaches in trustworthiness improvement, achieving the highest increase in accurate fairness (7.2\%) and individual fairness (9.7\%) across various sensitive attributes without compromising accuracy.
\end{abstract}

\section{Introduction}

Deep neural networks (DNNs) have demonstrated impressive performance in decision-making tasks. However, the widespread deployments of DNNs have increasingly raised public concerns about their trustworthiness \cite{Adv_E2, IF1, OOD1}. Studies have shown that DNNs are susceptible to various minor adversarial perturbations \cite{ R_IF1, R_IF2}, which can undermine not only the accuracy of predictions \cite{Adv2} but also the fairness of decisions made for similar individuals \cite{ADF}. Therefore, it is crucial to evaluate the impact of adversarial perturbations comprehensively, particularly in cases where robustness and fairness issues intertwine and may result in unacceptable defects. For instance, in civil domains such as income prediction \cite{banking}, loan approval \cite{adult}, and criminal justice risk assessment \cite{COMPAS}, accurate but biased adversarial predictions can exacerbate and perpetuate biases against specific individuals or groups. This lead to violations of anti-discrimination laws and ethical principles. Conversely, false but fair adversarial predictions may reduce the effectiveness of DNNs, resulting in decreased profits or compromised societal security. Hence, it is a formidable but essential challenge to ensure that DNNs remain both functionally and ethically correct when subjected to adversarial perturbations.

Accurate fairness \cite{AF} has recently been proposed to reconcile accuracy with individual fairness by enforcing that the predictions of an individual and its similar counterparts shall all conform to the individual's ground truth. An individual and its similar counterparts differ only in their sensitive features, such as gender, race, or age. Accurate fairness induces the notion of fairness confusion matrix, as shown in \tablename~\ref{Fairness Confusion Matrix}, which categorizes predictions as true fair, true biased, false fair, and false biased. \emph{True fair} (TF) predictions satisfy exactly the accurate fairness criterion.

\begin{itemize}
    \item Instances with either \emph{true biased} (TB) or \emph{false biased} (FB) predictions indicate that these instances and their similar counterparts receive different predictions, regardless of whether accurate or not.  These biased predictions are the targets of individual fairness evaluation.
    \item Instances with either \emph{false fair} (FF) or \emph{false biased} (FB) predictions indicate that these instances receive false predictions, no matter whether fair or biased. These false predictions are the primary concern of robustness evaluation. 
\end{itemize}

\begin{table}[!hbt]
\centering
  \caption{Fairness Confusion Matrix}
  \label{Fairness Confusion Matrix}
    \begin{tabular}{|c|cc|} \hline
    \diagbox{Accuracy}{Fairness} & Fair & Biased \\ \hline
    True & True Fair (TF) & True Biased (TB)\\ 
    False & False Fair (FF) & False Biased (FB)\\ \hline
    \end{tabular}
\end{table}

It is evident from Table \ref{Fairness Confusion Matrix} that both robustness and individual fairness evaluation approaches consistently overlook certain types of false or biased instances (TB or FF), when assessing the reliability of DNNs against false or biased adversarial perturbations. In this paper, we introduce RobustFair, a novel approach for evaluating the accurate fairness of DNNs under the influence of false or biased adversarial perturbations.

RobustFair leverages the total derivative to quantify loss function variations in response to input perturbations and identifies pertinent input features for perturbations through guidance from the fairness confusion matrix. These perturbations can have a dual impact on the instances and their similar counterparts: they may either elevate the loss functions of instances, thereby compromising robustness, or cause distinctive changes in loss functions of instances and their similar counterparts, thereby influencing individual fairness. Subsequently, RobustFair leverages the total derivative to approximate the loss function values for the generated adversarial instances and infer their ground truth based on these approximations. RobustFair offers a comprehensive approach for evaluating the accurate fairness of DNNs across various adversarial perturbation scenarios. It effectively addresses the limitations inherent in current robustness and individual fairness approaches, thus ensuring that DNNs maintain both functionally and ethically correct when confronted with intertwined issues of robustness and individual fairness.

We have implemented RobustFair and conducted an in-depth analysis of the fairness confusion matrix guided perturbations on several benchmark datasets, including Adult (Census Income) \cite{adult}, German Credit \cite{credit}, Bank Marketing \cite{banking}, and ProPublica Recidivism (COMPAS) \cite{COMPAS}. Our experimental results and quantile regression analyses demonstrate that the fairness confusion matrix could effectively direct the generation of false or biased adversarial instances. Compared to state-of-the-art techniques for testing robustness or individual fairness, RobustFair identifies significantly more false instances (3.365-8.183 times) and biased instances (2.217-7.818 times), as well as numerous false or biased instances (4.807-15.885 times) that were previously overlooked by the other approaches.

RobustFair further proposes a data augmentation strategy to utilize the generated adversarial instances for trustworthiness improvement. It employs the cosine similarity metric to assess the resemblance between these generated adversarial instances and the original training set. Then it prioritizes adversarial instances that closely resemble the original training set for dataset augmentation and model retraining. The empirical case studies further show that the adversarial instances identified by RobustFair outperform those identified by the other testing approaches, promoting 7.2\% accurate fairness and 9.7\% individual fairness on multiple sensitive attributes, without compromising accuracy

The main contributions of this paper are as follows:
\begin{itemize}
\item RobustFair provides a novel approach to evaluate the accurate fairness of DNNs against false or biased perturbations. It employs the total derivative to capture loss function variations by input perturbations, uses the fairness confusion matrix to guide the false or biased adversarial instance generation, and further infers the ground truth of these adversarial instances. This approach marks a significant advancement in comprehensively evaluating the trustworthiness of DNNs across various adversarial perturbation scenarios.

\item RobustFair applies a data augmentation strategy to leverage the generated adversarial instances for trustworthiness improvement, which helps to prioritize adversarial instances that closely resemble the original training data for augmenting the training dataset and retraining the original DNNs. As a result of this augmentation strategy, the retrained models demonstrate significant improvements in both accurate fairness and individual fairness without compromising accuracy.

\item Experimental results on fairness benchmark datasets demonstrate the effectiveness of RobustFair on false or biased perturbation evaluation. It outperforms the state-of-the-art robustness and individual fairness evaluation approaches in both the quantity and quality of adversarial instances identified. RobustFair can identify more adversarial instances with more diversity, which can be exploited to promote more accurate fairness, and individual fairness of DNNs, without sacrificing accuracy. This highlights the superiority of our fairness confusion directed adversarial evaluation.

\end{itemize}

The rest of this paper is organized as follows. Section 2 reviews briefly related work. Section 3 presents the RobustFair approach. Its evaluation results are reported and analyzed in Section 4. The paper is concluded in Section 5 with some future work.

\section{Related Work}

In this section, we briefly review relevant background, including Deep Neural Networks (DNNs), and gradient-based approaches for evaluating the robustness or fairness of DNNs.

\subsection{Deep Neural Networks} 

\begin{figure}[!htb]
    \centering
    \centerline{\includegraphics[width=0.4\linewidth]{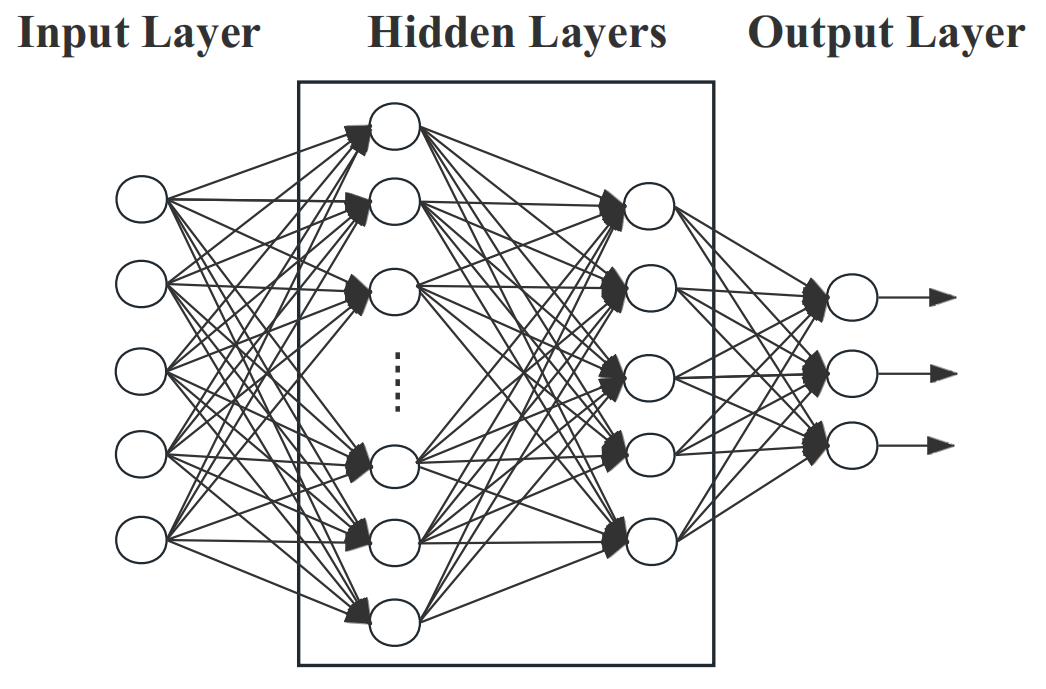}}
    \caption{The architecture of DNNs}
    \label{fig: DNNs Architecture}
\end{figure}

A Deep Neural Network (DNN) comprises an input layer, multiple hidden layers, and an output layer. The input layer receives external input, while the hidden layers process the output from the input layer or the preceding hidden layers for computations. The output layer receives the final output from the last hidden layer and generates the ultimate prediction. The architecture of DNNs is illustrated in \figureautorefname~\ref{fig: DNNs Architecture}.

In a DNN classifier, the input domains are denoted as $V \subseteq V_1 \times V_2 \times \cdots \times V_k$, where $V_h$ represents the domain of the $h$-th attribute, and the ground truth domain is denoted as $Y$. The DNN classifier, denoted as $f: V \rightarrow Y$, maps an input $v \in V $ to a prediction $\hat{y} = f(v)$. The DNN classifier consists of $L > 2$ layers, comprising one input layer, $L-2$ hidden layers (layers $2,\dots, L-1$), and one output layer (layer $L$). Let $s_p$ denote the number of neurons in the $p$-th layer, and $N^p = \{n_{p,q} |1 \leq q \leq s_p \}$ represents the set of neurons $n_{p,q}$ at layer $p$, where $1\leq p \leq L$.

For each neuron in the hidden layers or output layer, it calculates the weighted sum of the outputs of all neurons in its previous layer to obtain the output $o_{p,q}$ (as shown in Equation \eqref{output of a neuron}), where $w_{p-1,q,z}$ is the weight for the connection between $n_{p-1,z}$ and $n_{p,q}$, and $\psi$ is the activation function applied (e.g., Sigmoid, hyperbolic tangent (tanh), or rectified linear unit (ReLU) \cite{activation}).

\begin{equation}
\label{output of a neuron}
    o_{p,q}= \psi (\sum_{z=1}^{s_{p-1}} w_{p-1,q,z} o_{p-1,z})
\end{equation}

The output vector $\hat{y}$ of the DNN classifier consists of the activation values from the output layer, where each element of $\hat{y}$ represents the prediction probability for the corresponding class.

\subsection{Robustness and Fairness Evaluation for DNNs} 

\noindent\textbf{Robustness Evaluation} The concept of DNN robustness concerns about a DNN model's ability to withstand perturbations in model inputs without degrading its performance \cite{FGSM}. Typically, the robustness of a model is evaluated by measuring or specifically attacking its accuracy in the presence of adversarial perturbations \cite{R3}. The Fast Gradient Sign Method (FGSM) \cite{FGSM} and Projected Gradient Descent (PGD) attack \cite{PGD} are popular algorithms that may mislead the model by imperceptibly perturbing an input in the direction of the sign of the gradient of the loss function. The Auto Projected Gradient Descent (APGD) \cite{APGD}, a more advanced variant of PGD, considers iteratively the inertia direction of a search point with an adaptive step size. The Auto Conjugate Gradient (ACG) \cite{ACG} utilizes the conjugate gradient method to extensively search adversarial examples by updating a search point in more diverse directions. However, these approaches often prioritize robustness without adequately addressing individual fairness concerns. This focus on robustness may lead to models that perpetuate discrimination against specific individuals or groups if biased patterns exist in the training data.

\noindent\textbf{Fairness Evaluation}
Fairness asserts that DNNs treat similar groups or individuals equally. To evaluate group fairness violations, \cite{GTest1} proposes a statistical testing framework that uses optimal transport to project the empirical measure onto the set of group-fair probability models. \cite{GTest2} presents a statistical hypothesis testing to determine whether prediction errors are distributed similarly across similar groups. For individual fairness testing, the Adversarial Discrimination Finder (ADF) \cite{ADF, ADF1} utilizes a two-phase gradient search to enlarge prediction differences among similar individuals. The Efficient Individual Discriminatory Instances Generator (EIDIG) \cite{EIDIG} uses momentum optimization and intermediate gradient information to accelerate individual fairness testing. However, current individual fairness evaluation approaches often do not consider the impact of perturbations on prediction accuracy. This limitation arises because test oracles, such as ground truths, are typically unavailable for the generated biased instances. While this characteristic may be advantageous in certain scenarios, it also restricts the applicability of individual fairness evaluation, as evaluation outcomes may become intertwined with robustness issues. In an extreme scenario, enforcing identical predictions for all inputs would meet any fairness criterion but would severely compromise effectiveness.


The above gradient-based approaches evaluate DNNs against adversarial perturbations respectively from the perspectives of prediction accuracy and individual fairness. However, when DNNs encounter intertwined challenges related to robustness and individual fairness, current robustness and individual fairness evaluation approaches fail to ensure that DNNs maintain both functionally and ethically correct. The recently proposed criterion of accurate fairness establishes an orthogonal confusion aligning individual fairness with accuracy. In this paper, we introduce RobustFair, an accurate fairness evaluation approach that addresses the intertwined adversarial perturbation issues of robustness and individual fairness. It utilizes a customized two-stage gradient search for finer-grained false or biased prediction defects under the direction of the fairness confusion matrix.

\section{RobustFair}

In this section, we first formally describe the problem definition of accurate fairness evaluation against false or biased perturbations. We then detail how RobustFair solves the problem by generating instances that can undermine prediction accuracy (robustness) or introduce bias in predictions for similar inputs (individual fairness). Furthermore, we discuss the distinctions that RobustFair differs from current robustness or individual fairness testing approaches.

\subsection{Problem Definition}

Consider a finite and labeled dataset denoted as $V\subseteq X_1\times X_2\times \cdots X_m\times A_{1} \times A_{2}\times \cdots A_n$, where $X_i$ represents the domain of $i$-th non-sensitive attributes, $A_{j}$ represents the domain of $j$-th sensitive attributes, and $Y$ represents the ground truths. Each data point $v=(x_1,x_2,\dots,x_m,a_1,a_2,\dots a_n)\in V$ is associated with a ground truth value $y\in Y$. Let $I(v)\subseteq X_1\times X_2\times \cdots X_m\times A_{1} \times A_{2}\times \cdots A_n$ represent the set of similar instances to $v$, defined as instances that share the same values for non-sensitive attributes with $v$, formally denoted as $I(v)=\{(x_1,x_2,\dots,x_m,a_1',a_2',\dots a_n')~|~a_j'\in A_j, j \in [1,n]\}$. It is worth noting that $v\in I(v)$.

We define a classifier $f:V \rightarrow Y$ that is learned from training data. The prediction result of classifier $f$ for input $v$ is denoted as $\hat{y}=f(v)$. The term $l(y,f(v))$ refers to the loss function associated with the classifier's prediction $f(v)$ at the data point $v$. Additionally, $l(y,f(v'))$ represents a loss function associated with the prediction $f(v')$ for the similar counterpart $v'$  in $I(v)$ ($\exists a_j'\neq a_j, j\in [1,n]$).

We make the assumption that within a specific neighborhood of each point in $I(v)$, the partial derivatives of the loss function exist and that these partial derivative functions are continuous at those points.

The total derivative of the loss function at the data point $v$ and its similar counterpart $v'$ are expressed as follows:

\begin{equation}
\label{total derivative of the instance}
    \Delta l|_{v}= \sum_{i=1}^{m} \frac{\partial l}{\partial x_i} \Delta x_i + \sum_{j=1}^{n} \frac{\partial l}{\partial a_j} \Delta a_j
\end{equation}

\begin{equation}
\label{total derivative of the similar counterpart}
    \Delta l|_{v'}= \sum_{i=1}^{m} \frac{\partial l}{\partial x_i} \Delta x_i + \sum_{j=1}^{n} \frac{\partial l}{\partial a_j'} \Delta a_j'
\end{equation}

Within a specific neighborhood of the data point $v$, if the input feature $x_i$ undergoes a perturbation of $\Delta x_i$, the variation in the loss function can be linearly approximated as $\frac{\partial l}{\partial x_i}|_v \Delta x_i$, and the corresponding changes in the loss function for its similar counterpart can be expressed as $\frac{\partial l}{\partial x_i}|_{v'} \Delta x_i$.

Our objective is to evaluate the accurate fairness of DNNs by introducing perturbations to the input features, which have a dual effect on $\Delta l|_{v}$ in Equation \ref{total derivative of the instance} and $\Delta l|_{v'}$ in Equation \ref{total derivative of the similar counterpart}. These perturbations can either increase $\Delta l|_{v}$, thereby compromising robustness, or induce distinct changes between $\Delta l|_{v}$ and $\Delta l|_{v'}$, consequently influencing individual fairness.

\subsection{Workflow}
\begin{figure}[!htb]
    \centering
    \centerline{\includegraphics[width=\linewidth]{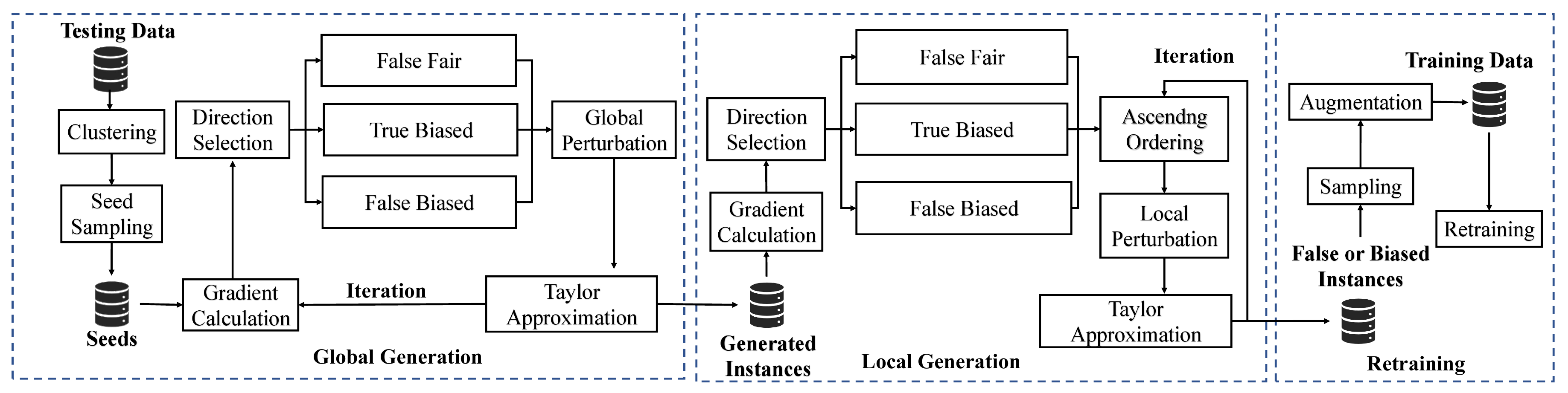}}
    \caption{The workflow of RobustFair}
    \label{fig: workflow}
\end{figure}

The high-level workflow of RobustFair is shown in \figureautorefname~\ref{fig: workflow}, which consists of two phases: global generation (Algorithm \ref{alg: Global_search}) and local generation (Algorithm \ref{alg: Local_search}).


The global generation phase searches for diverse instances to cover a broad range of an instance space. To scale the evaluation process, RobustFair starts by calling getSeeds($V$), which cluster the given testing data $V$ and sample randomly initial seeds uniformly from each cluster (Line 1). For each seed $(v,y)$, RobustFair selects its similar counterpart $v'\in I(v)$, with the largest absolute value of the gradient except for $v$, and then calculates the loss function gradients on seed $v$ and its similar counterpart $v'$ (Lines 6-8). These two gradients are examined to determine fairness confusion perturbations towards false fair (FF), true biased (TB), and false biased (FB) defects (Lines 9-11). For each directed perturbation $dir$, RobustFair then perturbs the seed $v$ accordingly with a fixed step size $p$, and approximates the ground truth $y_p$ for the resulting instance $v_p$ with total derivative groundTruth($v,y,g,f,v_p$) (Lines 13-14). The generated instance $(v_p,y_p)$ with its ground truth approximation is kept in set $R_T$ for the next iteration of global generation and in set $R_G$ for the subsequent local generation and evaluation (Lines 15-16).

In the local generation phase, RobustFair generates instances in the vicinity of each instance $(v,y)\in R_G$ generated in the global generation phase. RobustFair first determines the fairness confusion perturbations as in the global generation phase (Lines 4-9). Then it sorts each perturbation direction vector $dir$ in the ascending order of the absolute value of the derivative of each non-sensitive attribute. At Line 11, $min[i]$ represents the attribute index of the $i$-th least absolute derivative value. This allows RobustFair to select the attributes with the less absolute derivative values to perturb, which presumably results in fewer deviations between the prediction outputs. The minimal $iter$ absolute derivative values are selected one by one for minimal perturbations, with the ground truths approximated through the total derivative for the resulting instances (Lines 13-16). The generated instance $(v_p,y_p)$ with its ground truth approximation is kept in set $R_L$ for further evaluation (Line 17).

Finally, RobustFair utilizes the perturbed instances in $R_G\cup R_L$ to evaluate the accurate fairness of classifier $f$, with respect to their approximated ground truths. 

\begin{multicols}{2}
\begin{algorithm}[H]
\caption{Global Generation}
\label{alg: Global_search}
\textbf{Input:} model $f$, dataset $V$, search type $type$\\
\textbf{Output:} $R_G$
\begin{algorithmic}[1]
\STATE $R_G \gets \emptyset$; $S \gets $ getSeeds($V$);
\FOR{$i$ in range($iter$)}
    \STATE $R_T \gets \emptyset$;
    \FOR{each $(v,y) \in S$ with $v=(v)$}
        \STATE $D \gets \emptyset$;
        \STATE $v' \gets \mathop{\arg\max}\limits_{v' \in I(v),v'\neq v} \sum_{i=1}^{m} |\frac{\partial l}{\partial x_i}| + \sum_{j=1}^{n} |\frac{\partial l}{\partial a_j'}|$;
        \STATE $g \gets \partial \mathrm{loss}(y,f(v))/\partial v$;
        \STATE $g' \gets \partial \mathrm{loss}(y,f(v'))/\partial v'$;
        \STATE \textbf{Switch} $type$
            \STATE \quad \textbf{case} $False Fair$
                \STATE \quad  \quad $D\gets D\cup$\{dirFF($g,g'$)\};
            \STATE \quad  \textbf{case} $True Biased$
                \STATE \quad  \quad $D\gets D\cup$\{dirTB($g,g'$)\};
            \STATE \quad  \textbf{case} $False Biased$
                \STATE \quad  \quad $D\gets D\cup $\{dirFB($g,g'$)\};
        \FOR{each $dir \in D$}
            \STATE $v_p \gets v + p \cdot dir$;
            \STATE $y_p \gets$groundTruth($v,y,g,f,v_p$);
            \STATE $R_G \gets R_G \cup \{(v_p,y_p)\}$;
            \STATE $R_T \gets R_T \cup \{(v_p,y_p)\}$;
        \ENDFOR
    \ENDFOR
    \STATE $S \gets R_T$;
\ENDFOR
\STATE \textbf{return} $R_G$
\end{algorithmic}
\end{algorithm}

\begin{algorithm}[H]
\caption{Local Generation}
\label{alg: Local_search}
\textbf{Input}: model $f$ , search type $type$\\
\textbf{Output}:$R_L$ 
\begin{algorithmic}[1] 
\STATE $R_L$ $\gets\emptyset$;
\FOR{each $(v,y) \in R_G$ with $v=(v)$}
    \STATE $D \gets \emptyset$;
    \STATE $v' \gets \mathop{\arg\max}\limits_{v' \in I(v),v'\neq v} \sum_{i=1}^{m} |\frac{\partial l}{\partial x_i}| + \sum_{j=1}^{n} |\frac{\partial l}{\partial a_j'}|$;
    \STATE $g \gets \partial \mathrm{loss}(y,f(v))/\partial v$;
    \STATE $g' \gets \partial \mathrm{loss}(y,f(v'))/\partial v'$;
    \STATE \textbf{Switch} $type$
            \STATE \quad \textbf{case} $False Fair$
                \STATE \quad  \quad $D\gets D\cup$\{dirFF($g,g'$)\};
            \STATE \quad  \textbf{case} $True Biased$
                \STATE \quad  \quad $D\gets D\cup$\{dirTB($g,g'$)\};
            \STATE \quad  \textbf{case} $False Biased$
                \STATE \quad  \quad $D\gets D\cup $\{dirFB($g,g'$)\};
    \FOR {each $dir \in D$}
        \STATE $min\gets$ ascend($dir$);
        \FOR{$i$ in range($iter$)}
            \STATE $l_{dir}\gets$ zerosLike($dir$);
            \STATE $l_{dir}[min[i]]\gets dir[min[i]]$;
            \STATE $v_p\gets v+p*l_{dir}$;
            \STATE $y_p\gets$groundTruth($v,y,g,f,v_p$);
            \STATE $R_L \gets R_L \cup (v_p,y_p)$;
            \ENDFOR
    \ENDFOR
\ENDFOR
\STATE \textbf{return} $R_L$
\end{algorithmic}
\end{algorithm}
\end{multicols}

\subsection{Fairness Confusion Perturbations}

\begin{figure}[H]
    \centering
    \centerline{\includegraphics[width=0.60\linewidth,trim=10 0 10 0]{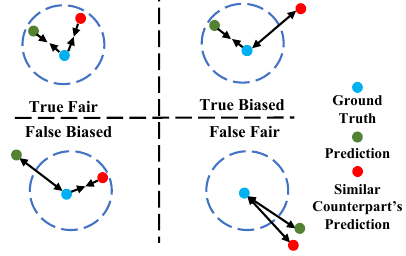}}
    \caption{Fairness Confusion Perturbations}
    \label{fig: Perturbation guidance}
\end{figure}

RobustFair aims at the accurate fairness defects categorized by the fairness confusion matrix under the alignment of accuracy and individual fairness. As depicted in \figureautorefname ~ \ref{fig: Perturbation guidance}, this figure illustrates the four categories of fairness confusion perturbations that relate to predictions for a perturbed instance $v_p$ and its similar counterpart $v_p' \in I(v_p)$ with $v_p'\neq v_p$, along with the instance's ground truth $y_p$.

If a perturbation results in a decrease in both $\Delta l|_{v}$ in Equation \eqref{total derivative of the instance} and $\Delta l|_{v'}$ in Equation \eqref{total derivative of the similar counterpart}, and the resulting predictions satisfy \emph{true fair} definition in accurate fairness \cite{AF}, i.e., $D(y_p, f(v_p))\leq Kd((v_p), (v_p))$ and $D(y_p, f(v_p'))\leq Kd((v_p), (v_p'))$, where $D(\cdot,\cdot)$ and $d(\cdot,\cdot)$ are distance metrics, and constant $K\geq 0$, then the perturbed instance is classified as \emph{true fair}. Such instances remain both accurate and fair when subjected to adversarial perturbations.

\begin{multicols}{2}

\begin{algorithm}[H]
\caption{False Fair Direction (dirFF) }
\label{alg: FF_search}
\textbf{Input}: gradients $g$,$g'$\\
\textbf{Output}: $dir$
\begin{algorithmic}[1] 
\STATE $dir \gets \text{zerosLike}(g)$
\FOR{each $g_i \in $ nonSensitive$(g)$}
    \IF{sign($g_i$)==sign($g'_i$)}
        \STATE $dir[i] \gets$ sign($g_i$)
    \ENDIF
\ENDFOR
\STATE \textbf{return} $dir$
\end{algorithmic}
\end{algorithm}

\end{multicols}

Otherwise, if the perturbation leads to $\Delta l|_{v}$ decrease in Equation \eqref{total derivative of the instance} and $\Delta l|_{v'}$ increase in Equation \eqref{total derivative of the similar counterpart}, and the resulting predictions are categorized as \emph{true biased}, then the perturbed instance is classified as \emph{true biased} adversarial instances. Conversely, if the perturbation leads to $\Delta l|_{v}$ increase in Equation \eqref{total derivative of the instance} and $\Delta l|_{v'}$ decrease in Equation \eqref{total derivative of the similar counterpart}, and the resulting predictions are categorized as \emph{fasle biased}, then the perturbed instance is classified as \emph{false biased} adversarial instances. Lastly, If a perturbation results in an increase in both $\Delta l|_{v}$ in Equation \eqref{total derivative of the instance} and $\Delta l|_{v'}$ in Equation \eqref{total derivative of the similar counterpart}, and the resulting predictions are categorized as \emph{fasle fair}, then the perturbed instance is classified as \emph{false fair} adversarial instances.

To identify these false or biased adversarial instances, we leverage the loss function gradients $g$ and $g'$ on an instance and its similar counterpart, as an intuitive and effective measure to direct perturbations on specific non-sensitive attributes in the loss function, Equation \eqref{total derivative of the instance} and \eqref{total derivative of the similar counterpart}. Given an instance, its similar counterpart that has the largest absolute value of the gradient except for $v$ is involved in gradient calculation, such that fairness confusion perturbations can be constructed from the perspective of the most diverse counterpart of the instance, promoting greater variety and inclusiveness in the RobustFair evaluation process. We propose Algorithms \ref{alg: FF_search}, \ref{alg: TB_search}, and \ref{alg: FB_search} that are devised for FF, TB, and FB perturbations, respectively. In these algorithms, nonSensitive($g$) returns the gradient component $g_i$ on each non-sensitive attribute, and $g_i'$ is the corresponding component of gradient $g'$. Let sign($g_i$) denote the sign function of $g_i$.

Algorithm~\ref{alg: FF_search} selects non-sensitive attributes with the same gradient signs to induce false fair perturbations (Line 3). Perturbing these attributes leads to consistent changes in the losses for both the instance and its similar counterpart. If sign($g_i$) is positive, a perturbation in the direction of $g_i$ can increase both $\Delta l|_{v}$ and $\Delta l|_{v'}$, causing the predictions $f(v_p)$ and $f(v_p')$ to deviate further from the ground truth $y_p$; otherwise, if sign($g_i$) is negative, a perturbation in the opposite direction of $g_i$ moves the predictions $f(v_p)$ and $f(v_p')$ away from the ground truth $y_p$ (Line 4).

In contrast, Algorithms~\ref{alg: TB_search} and \ref{alg: FB_search} select non-sensitive attributes with different gradient signs to induce true biased and false biased perturbations, respectively. Perturbing these attributes leads to opposite changes in the losses for the instance and its similar counterpart. In Algorithm~\ref{alg: TB_search} for a true biased perturbation, the prediction for the instance should get close to the ground truth ($\Delta l|_{v}$ decrease in Equation \eqref{total derivative of the instance}), while the one for its similar counterpart should deviate from the ground truth ($\Delta l|_{v'}$ increase in Equation \eqref{total derivative of the similar counterpart}). Thus, to enlarge the difference between the predictions, if the sign($g_i$) is not zero, a perturbation in the direction of $-$sign($g_i$) is determined to maneuver the prediction for the instance towards the ground truth (Line 5); otherwise, a perturbation in the direction of sign($g_i'$) is determined to maneuver the prediction for the similar counterpart away from the ground truth (Line 7). The opposite strategy applies for a false biased perturbation, as shown in Algorithm~\ref{alg: FB_search}.

\begin{multicols}{2}
\begin{algorithm}[H]
\caption{True Biased Direction (dirTB)}
\label{alg: TB_search}
\textbf{Input}: gradients $g$,$g'$\\
\textbf{Output}: $dir$
\begin{algorithmic}[1] 
\STATE $dir \gets \text{zerosLike}(g)$;
\FOR{each $g_i \in $ nonSensitive$(g)$}
    \IF{sign($g_i$) $\neq$ sign($g'_i$)}
        \IF{sign($g_i) \neq 0$}
            \STATE $dir[i]\gets -$ sign($g_i$);
        \ELSE
            \STATE $dir[i]\gets$ sign($g_i'$);
    \ENDIF
    \ENDIF
\ENDFOR
\STATE \textbf{return} $dir$
\end{algorithmic}
\end{algorithm}

\begin{algorithm}[H]
\caption{False Biased Direction (dirFB)}
\label{alg: FB_search}
\label{alg:find_direction}
\textbf{Input}: gradients $g$,$g'$ \\
\textbf{Output}: $dir$
\begin{algorithmic}[1] 
\STATE $dir \gets \text{zerosLike}(g)$;
\FOR{each $g_i \in $ nonSensitive$(g)$}
    \IF{sign($g_i$) $\neq$ sign($g'_i$)}
        \IF{sign($g_i) \neq 0$}
            \STATE $dir[i] \gets $ sign($g_i$)
        \ELSE
            \STATE $dir[i] \gets -$ sign($g_i'$)
        \ENDIF
    \ENDIF
\ENDFOR
\STATE \textbf{return} $dir$
\end{algorithmic}
\end{algorithm}
\end{multicols}

\subsection{Ground Truth Approximation}

Accurate fairness evaluation and fair confusion perturbations both rely on the ground truths of the generated instances, which however are generally not available. The total derivative of a function at a specific point offers the most accurate linear approximation of the function in the vicinity of that point concerning its input variables. RobustFair approximates the loss of the generated instance $v_p$ through the total derivative \cite{Mathematical_Analysis}, as depicted below:

\begin{equation}
l(y_p,f(v_p))-l(y,f(v)) \approx  \Delta l|_{v}
\end{equation}

\begin{equation}
\label{total derivative approximation}
l(y_p,f(v_p)) \approx l(y,f(v)) + \sum_{i=1}^{m} \frac{\partial l}{\partial x_i} \Delta x_i + \sum_{j=1}^{n} \frac{\partial l}{\partial a_j} \Delta a_j
\end{equation}


where $v_p$ is generated by perturbing instance $v$, $y$ is the ground truth of $v$, $l(\cdot,\cdot)$ is the loss function, and $ \frac{\partial l}{\partial x_i}$, $\frac{\partial l}{\partial a_j}$ are the derivative on $x_i,a_j$. Then, the approximated ground truth $y_p$ can be derived from Equation \eqref{total derivative approximation}. For the Mean Square Error ($\mathrm{MSE}$) loss function, groundTruth($v,y,g,f,v_p$) can be implemented as below.
\begin{equation}
\text{groundTruth}(v,y,g,f,v_p)=\left\{\begin{matrix}
y^- & |y^--y|<|y^+-y|\\
y^+ & \text{otherwise}
\end{matrix}\right.
\end{equation}
where $y^+=f(v_p)+\sqrt[2]{|\mathrm{MSE}(y,f(v))+ g(v_p-v)|}$ and $y^-=f(v_p)-\sqrt[2]{|\mathrm{MSE}(y,f(v))+g(v_p-v)|}$. That is, the approximated ground truth $y_p$ for $v_p$ is the one of $y^-$ and $y^+$ that is closer to $y$.

\subsection{Data Augmentation Strategy}

To enhance the trustworthiness of the original DNNs using the generated adversarial instances, RobustFair additionally utilizes the cosine similarity metric to quantify the similarity between the generated adversarial instances and the original training set. The generative similarity of an adversarial instance is defined as the maximum cosine similarity between the generated instance $(v_p, y_p)$ and a cluster centroid $(v_c, y_c)$ within the set $C$, which is established through the clustering function $getSeeds(V)$ during the global generation process:

\begin{equation}
\label{cosine similarity}
S_{(v_p,y_p)} =\mathop{\arg\max}\limits_{(v_c,y_c) \in C}  cos((v_p,y_p),(v_c,y_c))
\end{equation}

RobustFair prioritizes adversarial instances that closely resemble the original training set ($S_{(v_p,y_p)}>\sigma$) for augmenting the original training dataset. Here, $\sigma$ represents a threshold used to select the adversarial instances that are sufficiently similar to the original training set. After this selection process, RobustFair proceeds to retrain the original DNNs using the augmented dataset to enhance accurate fairness.

\subsection{Distinctions Analysis}

Finally, we highlight the distinctions between RobustFair and current robustness and individual fairness evaluation approaches.

\noindent\textbf{Multiple-Perspective Evaluation.}
Existing evaluation approaches often focus on a single aspect of DNNs, such as robustness or individual fairness. However, when DNNs encounter intertwined challenges related to both robustness and individual fairness, current evaluation approaches in either category fail to ensure that DNNs maintain both functional correctness and ethical fairness. In contrast, our approach, RobustFair, offers a comprehensive evaluation methodology that assesses the accurate fairness of DNNs from two distinct perspectives: robustness and individual fairness. A model with high accurate fairness is expected to be less affected by false or biased adversarial perturbations, thus effectively addressing these critical concerns and ensuring both functional correctness and ethical fairness.

Evaluating the accurate fairness of DNNs from multiple perspectives also enhances the diversity of gradient search. 

\noindent\textbf{Multi-Direction for Robustness Evaluation.}
In typical robustness evaluations, techniques such as \cite{FGSM, PGD, APGD, ACG} utilize the gradient or conjugate gradient \cite{ACG} of the instance to determine the perturbation direction. In contrast, our approach employs the total derivative of the instance and its similar counterparts, considering individual fairness when identifying the robustness perturbation direction. This allows us to perturb attributes that could lead to false biased or false fair predictions. RobustFair introduces two perturbation directions for robustness evaluation. 

\noindent\textbf{Multi-Direction for Individual Fairness Evaluation.}
Individual fairness evaluations aim to increase the prediction distance between instances and their similar counterparts. However, the gradient-based perturbation strategy adopted by \cite{ADF, EIDIG} is quite similar to our False Fair Direction \ref{alg: FF_search}, which generates false fair adversarial instances by increasing the loss functions or predictions. In contrast, our approach considers the ground truth as an intermediate node when identifying the individual fairness perturbation direction. This allows us to perturb the attributes that would generate true biased or false biased predictions, as instances or similar counterparts either approach or move away from the ground truth after perturbation. RobustFair introduces two perturbation directions for individual fairness evaluation.

\noindent \textbf{Ground Truth Approximation}. Current evaluation methods lack the capability to estimate the ground truth for perturbed instances. When instance attributes are perturbed, the ground truth of these instances may also change. In robustness evaluation approaches, the ground truth of the initial input instance is typically assumed as the ground truth for the perturbed instance. In contrast, individual fairness evaluation approaches do not consider the ground truth of the perturbed instance. Our method stands out by utilizing the total derivative of the loss function to approximate the loss functions for the perturbed instances and then infer the ground truth based on these approximations. This approach offers a means to estimate the ground truth of the perturbed instance. To the best of our knowledge, this is the first method to offer a way to estimate the ground truth of a perturbed instance after perturbation, addressing a significant gap in existing approaches.


\section{Experimental Evaluation}

We implement RobustFair in Python 3.8 with TensorFlow 2.4.1. The implementation is evaluated on a Ubuntu 18.04.3 system, equipped with Intel Xeon Gold 6154 @3.00GHz CPUs, GeForce RTX 2080 Ti GPUs, and 512GB memory. The source code, experimental datasets, and models are submitted in the supplementary material and will be made publicly available later.

\subsection{Experiment Setup}

Four popular fairness datasets Adult (Census Income) \cite{adult}, German Credit \cite{credit}, Bank Marketing \cite{banking}, and ProPublica Recidivism (COMPAS) \cite{COMPAS} are considered for experimental evaluation. The sizes, sensitive attributes, and the baseline models (BL) trained with these datasets are reported in Table~\ref{tab-datasets}, where $A$($k$) indicates that attribute $A$ has $k$ values, and FCNN($l$) denotes a fully connected neural network classifier with $l$ layers. Specifically, we consider all the sensitive attributes jointly, instead of separately, to conduct complex case studies.

\begin{table}[!htb]
 \centering
 \caption{Datasets and Models}
\label{tab-datasets}
\resizebox{0.75\columnwidth}{!}{
\begin{tabular}{|l|c|c|l|}
\hline
\multicolumn{1}{|c|}{Dataset} & Size & Model & Sensitive Attributes \\ \hline
\makecell*[l]{German Credit} & 1000 & \makecell*[c]{FCNN(6)} & \makecell*[l]{gender(2), age(51)} \\ \hline
\makecell*[l]{Bank Marketing} & 45211 & \makecell*[c]{FCNN(6)} & \makecell*[l]{age(77)} \\ \hline
\makecell*[l]{Adult (Census Income)} & 45222 & \makecell*[c]{FCNN(6)}&\makecell*[l]{gender(2), age(71), race(5)} \\ \hline
\makecell*[l]{ProPublica Recidivism (COMPAS)} & 6172 & \makecell*[c]{FCNN(6)} & \makecell*[l]{gender(2), age(71), race(6)} \\ \hline
\end{tabular}
}
\end{table}

The state-of-the-art robustness testing techniques, including Fast Gradient Sign Method (FGSM) \cite{FGSM}, Projected Gradient Descent (PGD) attack \cite{PGD}, Auto Projected Gradient Descent (APGD) \cite{APGD}, and Auto Conjugate Gradient (ACG) \cite{ACG}, and individual fairness testing techniques, including Adversarial Discrimination Finder (ADF) \cite{ADF1} and Efficient Individual Discriminatory Instances Generator (EIDIG) \cite{EIDIG} are involved for comparison. 

\subsection{Results and Discussion}

The objective of our experiments is to address the following five inquiries.

\textit{Q1: How efficient are fairness confusion perturbations?}

\begin{figure}[!htb]
\centering
    \begin{minipage}[t]{0.30\linewidth}
        \centering
        \includegraphics[width=1.0\linewidth]{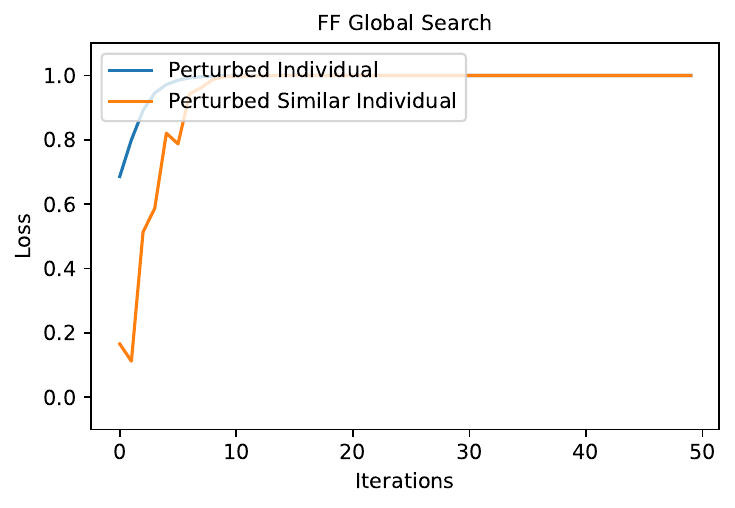}
    \end{minipage}
    \begin{minipage}[t]{0.30\linewidth}
        \centering
        \includegraphics[width=1.0\linewidth]{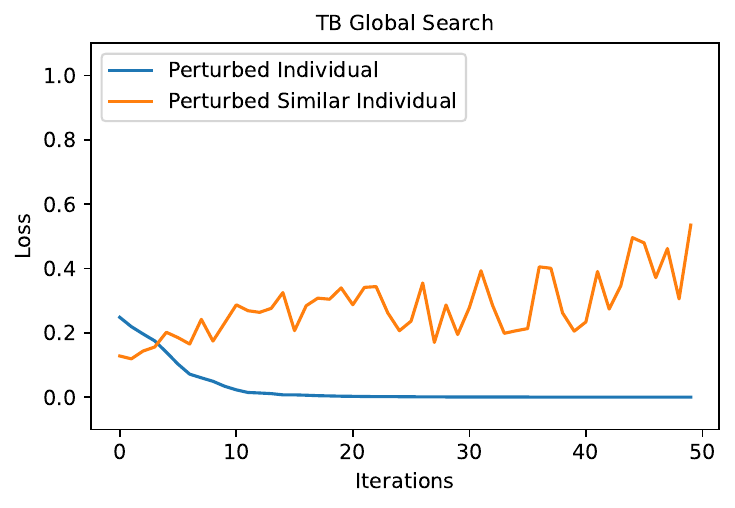}
    \end{minipage}
    \begin{minipage}[t]{0.30\linewidth}
        \centering
        \includegraphics[width=1.0\linewidth]{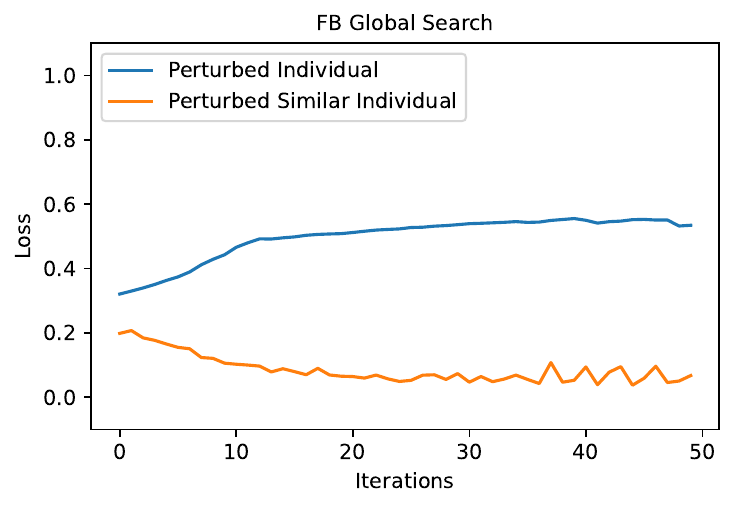}
    \end{minipage}
    \caption{Global Search Process on Bank Marketing Dataset}
    \label{bank G search}
\end{figure}

\begin{figure}[!htb]
\centering
    \begin{minipage}[t]{0.30\linewidth}
        \centering
        \includegraphics[width=1.0\linewidth]{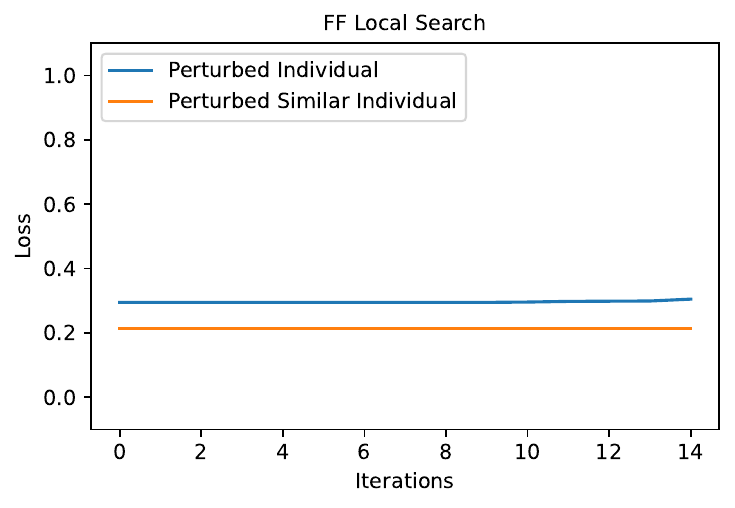}
    \end{minipage}
    \begin{minipage}[t]{0.30\linewidth}
        \centering
        \includegraphics[width=1.0\linewidth]{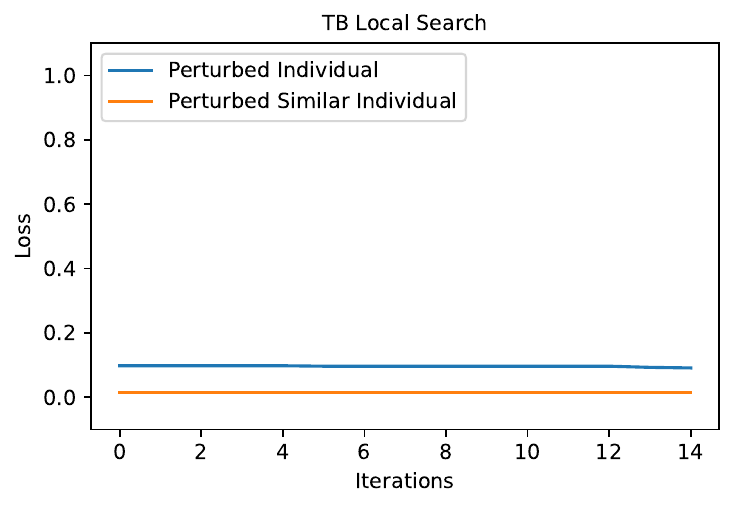}
    \end{minipage}
    \begin{minipage}[t]{0.30\linewidth}
        \centering
        \includegraphics[width=1.0\linewidth]{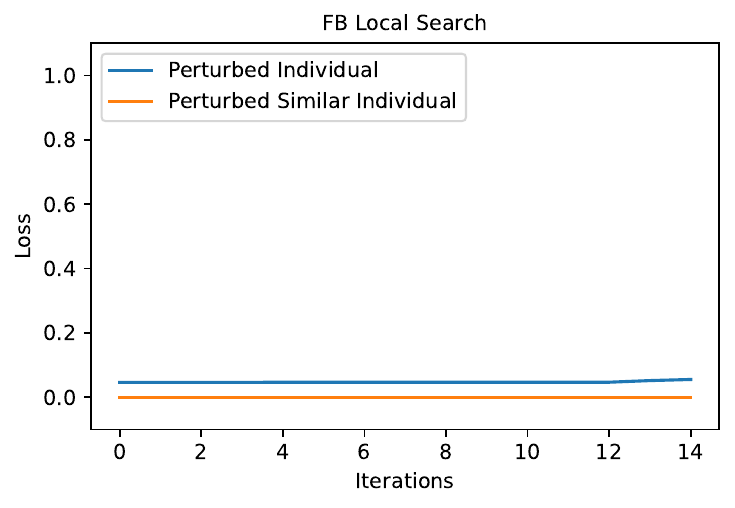}
    \end{minipage}

    \caption{Local Search Process on Bank Marketing Dataset}
    \label{bank L search}
\end{figure}

Figures \ref{bank G search} and \ref{bank L search} illustrate the trends in the loss function for an instance and its similar counterpart during the global generation and local generation processes of fairness confusion perturbations (false fair, true bias, and false bias). These figures are generated by monitoring the loss function values of an instance in the Bank Marketing dataset when subjected to false or biased perturbations, with the perturbation step size $p=0.01$ and iteration $iter=50$.

As illustrated in Figure \ref{bank G search}, during the false fair global generation process, the false fair perturbations yield a similar impact on both the instance and its similar counterparts: their loss functions all exhibit an increase. This indicates that the perturbed instance may receive a false treatment according to its ground truth but remain equal to its counterpart.

In contrast, during true biased and false biased global generation, adversarial perturbations have a distinct impact on the instance and its similar counterparts. One experiences an increase while the other undergoes a decrease, resulting in the perturbed instance receiving differential treatment compared to its counterpart. Specifically, in True Biased global generation, the instance's loss function decreases while its similar counterpart's loss function increases. Conversely, in False Biased global generation, these loss functions exhibit opposite changes.

Figure \ref{bank L search} illustrates the changes in the loss function during the local generation process. It is apparent that the local perturbations have minimal effects on the instance and its similar counterpart. This is because the objective of the local process is to identify false or biased adversarial instances in the vicinity of the input instance. In this process, we arrange the non-sensitive attributes in descending order based on their absolute partial derivative values and sequentially perturb these sorted non-sensitive attributes. These perturbations are also guided by the fairness confusion matrix, with the goal of minimizing changes to the accurate fairness of the predictions.

We also notice the number of iterations in the local generation process is less than 50. The local iteration number is determined by selecting the lower value between the number of non-sensitive attributes and the specified iteration parameters. In the Bank Marketing dataset, which encompasses 16-dimensional attributes, including 1 sensitive attribute (age), the maximum number of iterations for local generation is limited to 15.

Additionally, we employ quantile regression \cite{quantile_regression} to analyze the trends in the loss function of 100 instances in the Bank Marketing Dataset. Initially, we apply K-Means clustering \cite{KMeans} to the test dataset and carefully select 100 initial seeds from the resulting clusters. Subsequently, we record the loss function values of these instances when subjected to the global generation and local generation processes of fairness confusion perturbations.
Regarding the loss function as a random variable given the number of iterations, we employ quantile regression to model the median value of the loss function $l$ as a linear function of the number of iterations $iter$, as represented by Equation (\ref{reg}):


\begin{equation}
	\label{reg}
	\mathbb{Q} _\frac{1}{2}\left( l|iter \right)= \beta_0 +\beta_1\cdot iter
\end{equation}

Here, $\mathbb{Q}_{\frac{1}{2}}(l | \text{iter})$ represents the median of the loss function given the number of iterations, and $\beta=[\beta_0, \beta_1]^T$ is the parameter vector estimated through the optimization problem expressed in Equation \ref{beta parameter}:

\begin{equation}
 \label{beta parameter}
	\hat{\beta}=\underset{\beta \in R^2}{\mathrm{arg}\min}\sum_{k=1}^n{\left| l_k-\beta_0-\beta_1 \cdot iter_k \right|}
\end{equation}

The sign of $\hat{\beta_1}$ indicates the trend of the loss function with respect to the number of iterations. If $\text{sign}(\hat{\beta_1}) = 1$, it signifies that the median of the loss function increases as the number of iterations grows, indicating an upward trend. Conversely, if $\text{sign}(\hat{\beta_1}) = -1$, it indicates that the median of the loss function decreases as the number of iterations increases, reflecting a downward trend. In the case where $\text{sign}(\hat{\beta_1}) = 0$, it implies that the median of the loss function remains relatively constant as the number of iterations increases.

\begin{figure}[!htb]
\centering
    \begin{minipage}[t]{0.30\linewidth}
        \centering
        \includegraphics[width=1.0\linewidth]{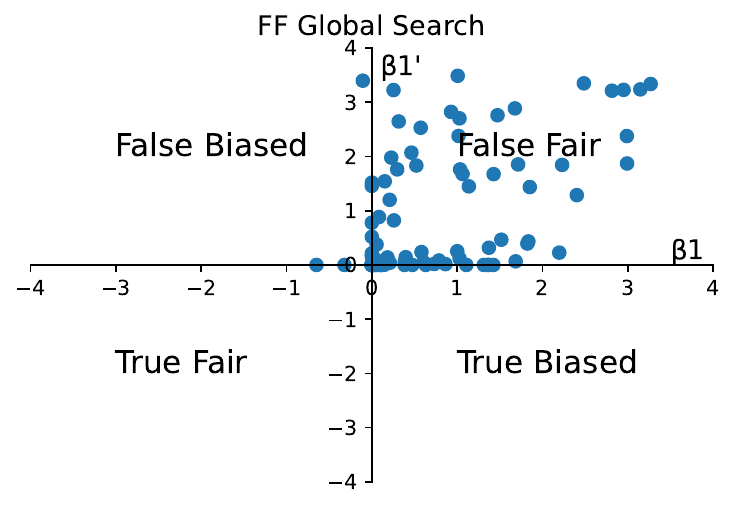}
    \end{minipage}
    \begin{minipage}[t]{0.30\linewidth}
        \centering
        \includegraphics[width=1.0\linewidth]{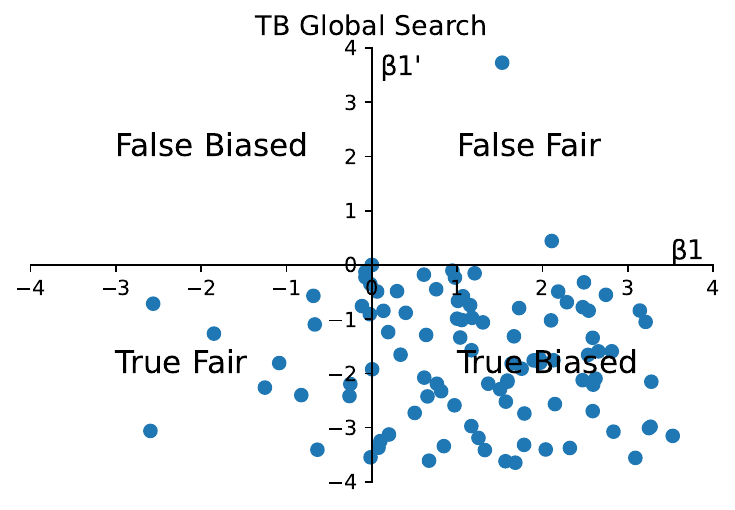}
    \end{minipage}
    \begin{minipage}[t]{0.30\linewidth}
        \centering
        \includegraphics[width=1.0\linewidth]{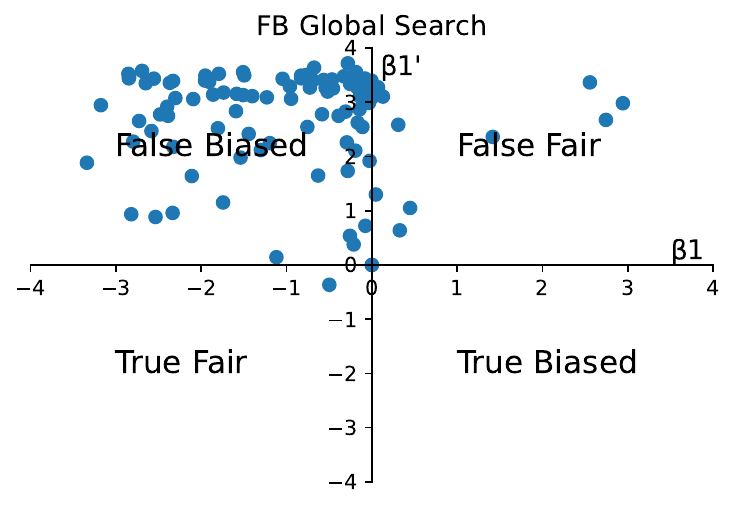}
    \end{minipage}

    \caption{Global Fairness Confusion Perturbation Analysis on Bank dataset}
    \label{Bank global analysis}
\end{figure}

\begin{figure}[!htb]
\centering
    \begin{minipage}[t]{0.30\linewidth}
        \centering
        \includegraphics[width=1.0\linewidth]{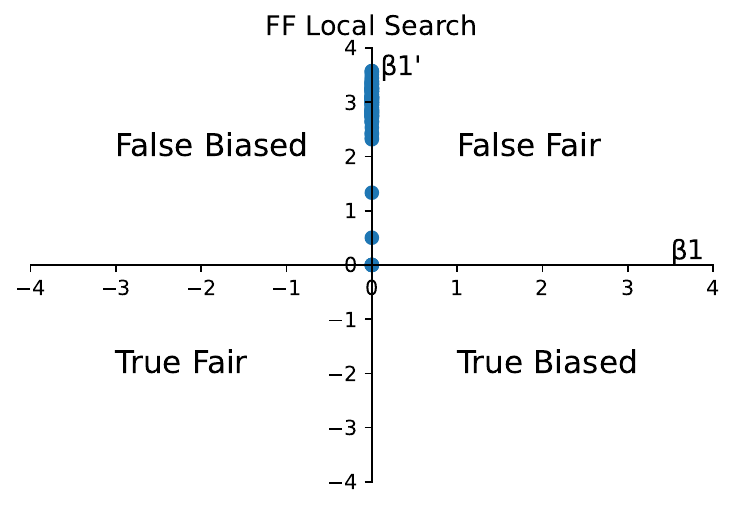}
    \end{minipage}
    \begin{minipage}[t]{0.30\linewidth}
        \centering
        \includegraphics[width=1.0\linewidth]{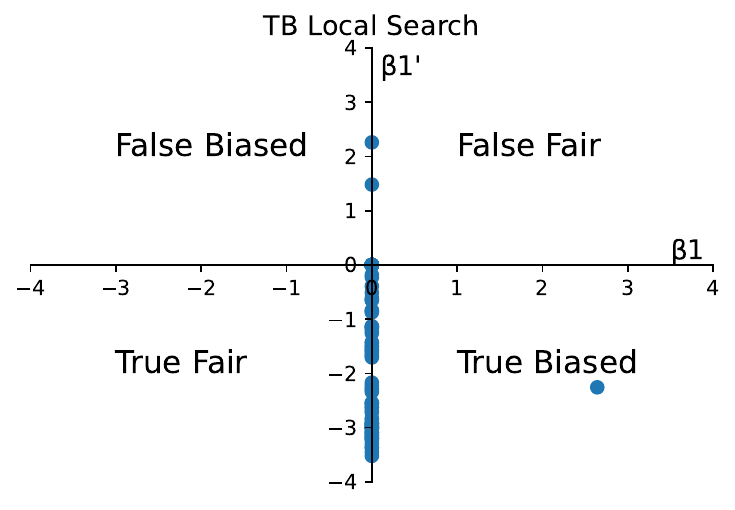}
    \end{minipage}
    \begin{minipage}[t]{0.30\linewidth}
        \centering
        \includegraphics[width=1.0\linewidth]{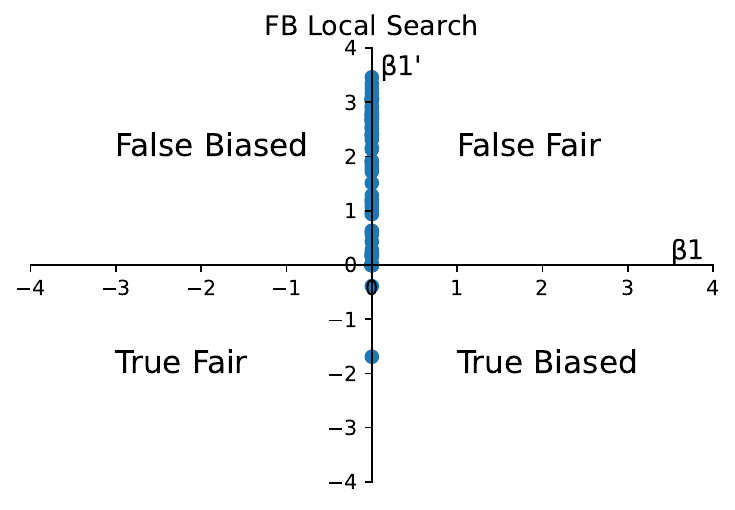}
    \end{minipage}

    \caption{Local Fairness Confusion Perturbation Analysis on Bank dataset}
    \label{Bank local analysis}
\end{figure}

The statistical results of quantile regression are presented in \figureautorefname~\ref{Bank global analysis} and \ref{Bank local analysis}. We use the parameters $\hat{\beta_1}$ and $\hat{\beta_1'}$ derived from the instances and their similar counterparts to categorize the coordinate system into four areas, namely: \textit{True Fair} (sign$(\hat{\beta_1})$ = -1, sign$(\hat{\beta_1'})$ = -1), \textit{True Biased} (sign$(\hat{\beta_1})$ = -1, sign$(\hat{\beta_1'})$ = 1), \textit{False Fair} (sign$(\hat{\beta_1})$ = 1, sign$(\hat{\beta_1'})$ = 1), and \textit{False Biased} (sign$(\hat{\beta_1})$ = 1, sign$(\hat{\beta_1'})$ = -1).


As depicted in \figureautorefname~\ref{Bank global analysis}, most instances following false fair, true biased, and false biased global perturbations are situated in their respective categories of false fair, true biased, and false biased areas. Turning to the local perturbation results in \figureautorefname~\ref{Bank local analysis}, the majority of perturbed instances are positioned along the ordinate axis with $sign(\hat{\beta_1})=0$. This suggests that local perturbations have minimal impact on the instances themselves but significantly affect their similar counterparts more, thereby searching for false or biased adversarial instances in the vicinity of the input instances.

Due to the page limit, we herein discuss the fairness confusion perturbations on the Bank Marketing dataset. Similar observations can be made on the German Credit, Adult, and ProPublica Recidivism datasets. Please refer to the supplementary material for the detailed experimental results.

\textit{Answer to Q1: Fairness confusion perturbations can generate false or biased adversarial perturbations that produce a dual impact on instances and their similar counterparts: they can either elevate loss functions of instances, compromising robustness, or induce distinctive changes in loss functions of instances and their similar counterparts, thereby influencing individual fairness.}

\textit{Q2: How efficient is RobustFair in trustworthiness evaluation?}

Table~\ref{Trustworthiness Statistics} displays the average statistics obtained from five runs of seven trustworthiness evaluation techniques applied to the baseline model. These evaluations are conducted across four fairness datasets: Adult, German Credit, Bank Marketing, and ProPublica Recidivism. 


\begin{table}[!htb]
\caption{Statistics of Techniques across Four Fairness Datasets}
\label{Trustworthiness Statistics}
\resizebox{1.0\linewidth}{!}{
    \begin{tabular}{|c|c|ll|ll|l|l|l|}
    \hline
    Model & Dataset & \multicolumn{1}{c|}{$R_f$} & \multicolumn{1}{c|}{$N_f$} & \multicolumn{1}{c|}{$R_b$} & \multicolumn{1}{c|}{$N_b$} & \multicolumn{1}{c|}{$R_{f|b}$} & \multicolumn{1}{c|}{$N_{f|b}$} & \multicolumn{1}{c|}{$sum$} \\ \hline \hline
    FGSM & \multirow{7}{*}{Adult} & \multicolumn{1}{c|}{17.2\%} & 17.000 & \multicolumn{2}{c|}{\multirow{4}{*}{——}} & 17.2\% & 17.000 & 99.000 \\ \cline{1-1} \cline{3-4} \cline{7-9} 
    PGD &  & \multicolumn{1}{c|}{27.6\%} & 67.000 & \multicolumn{2}{c|}{} & 27.6\% & 67.000 & 243.000 \\ \cline{1-1} \cline{3-4} \cline{7-9} 
    APGD &  & \multicolumn{1}{c|}{\textbf{59.7\%}} & 3019.000 & \multicolumn{2}{c|}{} & 59.7\% & 3019.000 & 5056.000 \\ \cline{1-1} \cline{3-4} \cline{7-9} 
    ACG &  & \multicolumn{1}{c|}{45.0\%} & 2159.000 & \multicolumn{2}{c|}{} & 45.0\% & 2159.000 & 4798.000 \\ \cline{1-1} \cline{3-9} 
    ADF &  & \multicolumn{2}{c|}{\multirow{2}{*}{——}} & \multicolumn{1}{c|}{\textbf{88.4\%}} & 2091.400 & 88.4\% & 2091.400 & 2366.200 \\ \cline{1-1} \cline{5-9} 
    EIDIG &  & \multicolumn{2}{c|}{} & \multicolumn{1}{c|}{33.9\%} & 451.000 & 33.9\% & 451.000 & 1326.200 \\ \cline{1-1} \cline{3-9} 
    RobustFair &  & \multicolumn{1}{c|}{51.7\%} & \textbf{7347.600} & \multicolumn{1}{c|}{66.7\%} & \textbf{9478.200} & \multicolumn{1}{c|}{\textbf{90.8\%}} & \multicolumn{1}{c|}{\textbf{12895.800}} & \textbf{14209.200} \\ \hline \hline
    FGSM & \multirow{7}{*}{\begin{tabular}[c]{@{}c@{}}Bank\\Marketing\end{tabular}} & \multicolumn{1}{c|}{13.0\%} & 13.000 & \multicolumn{2}{c|}{\multirow{4}{*}{——}} & 13.0\% & 13.000 & 100.000 \\ \cline{1-1} \cline{3-4} \cline{7-9} 
    PGD &  & \multicolumn{1}{c|}{17.8\%} & 65.000 & \multicolumn{2}{c|}{} & 17.8\% & 65.000 & 365.000 \\ \cline{1-1} \cline{3-4} \cline{7-9} 
    APGD &  & \multicolumn{1}{c|}{36.7\%} & 3538.000 & \multicolumn{2}{c|}{} & 36.7\% & 3538.000 & 9645.000 \\ \cline{1-1} \cline{3-4} \cline{7-9} 
    ACG &  & \multicolumn{1}{c|}{21.3\%} & 1956.000 & \multicolumn{2}{c|}{} & 21.3\% & 1956.000 & 9169.000 \\ \cline{1-1} \cline{3-9} 
    ADF &  & \multicolumn{2}{c|}{\multirow{2}{*}{——}} & \multicolumn{1}{c|}{\textbf{81.3\%}} & 3652.800 & 81.3\% & 3652.800 & 4494.400 \\ \cline{1-1} \cline{5-9} 
    EIDIG &  & \multicolumn{2}{c|}{} & \multicolumn{1}{c|}{22.3\%} & 628.200 & 22.3\% & 628.200 & 2811.800 \\ \cline{1-1} \cline{3-9} 
    RobustFair &  & \multicolumn{1}{c|}{\textbf{46.9\%}} & \textbf{4688.000} & \multicolumn{1}{c|}{47.5\%} & \textbf{4746.000} & \multicolumn{1}{c|}{\textbf{78.9\%}} & \multicolumn{1}{c|}{\textbf{7893.600}} & \textbf{10000.800} \\ \hline \hline
    FGSM & \multirow{7}{*}{COMPAS} & \multicolumn{1}{c|}{3.1\%} & 3.000 & \multicolumn{2}{c|}{\multirow{4}{*}{——}} & 3.1\% & 3.000 & 98.000 \\ \cline{1-1} \cline{3-4} \cline{7-9} 
    PGD &  & \multicolumn{1}{c|}{4.6\%} & 8.000 & \multicolumn{2}{c|}{} & 4.6\% & 8.000 & 174.000 \\ \cline{1-1} \cline{3-4} \cline{7-9} 
    APGD &  & \multicolumn{1}{c|}{6.4\%} & 307.000 & \multicolumn{2}{c|}{} & 6.4\% & 307.000 & 4783.000 \\ \cline{1-1} \cline{3-4} \cline{7-9} 
    ACG &  & \multicolumn{1}{c|}{6.1\%} & 287.000 & \multicolumn{2}{c|}{} & 6.1\% & 287.000 & 4716.000 \\ \cline{1-1} \cline{3-9} 
    ADF &  & \multicolumn{2}{c|}{\multirow{2}{*}{——}} & \multicolumn{1}{c|}{59.1\%} & 1211.400 & 59.1\% & 1211.400 & 2048.400 \\ \cline{1-1} \cline{5-9} 
    EIDIG &  & \multicolumn{2}{c|}{} & \multicolumn{1}{c|}{27.4\%} & 394.400 & 27.4\% & 394.400 & 1414.800 \\ \cline{1-1} \cline{3-9} 
    RobustFair &  & \multicolumn{1}{c|}{\textbf{11.7\%}} & \textbf{903.400} & \multicolumn{1}{c|}{\textbf{67.5\%}} & \textbf{5214.000} & \multicolumn{1}{c|}{\textbf{75.8\%}} & \multicolumn{1}{c|}{\textbf{5853.000}} & \textbf{7721.400} \\ \hline \hline
    FGSM & \multirow{7}{*}{\begin{tabular}[c]{@{}c@{}}German\\Credit\end{tabular}} & \multicolumn{1}{c|}{19.0\%} & 15.000 & \multicolumn{2}{c|}{\multirow{4}{*}{——}} & 19.0\% & 15.000 & 79.000 \\ \cline{1-1} \cline{3-4} \cline{7-9} 
    PGD &  & \multicolumn{1}{c|}{22.0\%} & 24.000 & \multicolumn{2}{c|}{} & 22.0\% & 24.000 & 109.000 \\ \cline{1-1} \cline{3-4} \cline{7-9} 
    APGD &  & \multicolumn{1}{c|}{\textbf{53.1\%}} & 909.000 & \multicolumn{2}{c|}{} & 53.1\% & 909.000 & 1712.000 \\ \cline{1-1} \cline{3-4} \cline{7-9} 
    ACG &  & \multicolumn{1}{c|}{39.9\%} & 633.000 & \multicolumn{2}{c|}{} & 39.9\% & 633.000 & 1585.000 \\ \cline{1-1} \cline{3-9} 
    ADF &  & \multicolumn{2}{c|}{\multirow{2}{*}{\textbf{——}}} & \multicolumn{1}{c|}{\textbf{86.7\%}} & 909.200 & 86.7\% & 909.200 & 1047.800 \\ \cline{1-1} \cline{5-9} 
    EIDIG &  & \multicolumn{2}{c|}{} & \multicolumn{1}{c|}{34.2\%} & 308.400 & 34.2\% & 308.400 & 901.400 \\ \cline{1-1} \cline{3-9} 
    RobustFair &  & \multicolumn{1}{c|}{51.3\%} & \textbf{3234.400} & \multicolumn{1}{c|}{75.5\%} & \textbf{4759.800} & \multicolumn{1}{c|}{\textbf{95.7\%}} & \multicolumn{1}{c|}{\textbf{6029.400}} & \textbf{6303.000} \\ \hline
    \end{tabular}
        
}
\end{table}

To prepare the comparison, we utilized K-Means clustering \cite{KMeans} to the test dataset and selected 100 initial seeds from the resulting clusters for each testing technique. For ADF, EIDIG, and RobustFair, we perform 10 global and local iterations. In contrast, PGD, APGD, and ACG use 100 iterations as they do not involve local searches. For FGSM, 100 initial seeds are directly employed, as it does not require iteration searches. It is worth noting that we ensure each technique conducts an equal number of searches (10*10) for each seed, except for FGSM.

Columns $N_f$, $N_b$, and $N_{f|b}$, or $R_f$, $R_b$, and $R_{f|b}$, respectively, display the counts or percentages of false, biased, and false or biased adversarial instances detected by each technique. Higher counts or percentages indicate better efficiency for evaluating the robustness, individual fairness, or accurate fairness of a DNNs. The columns labeled $sum$ represent the number of generated instances.

It is evident that robustness testing approaches such as FGSM \cite{FGSM}, PGD \cite{PGD}, APGD \cite{APGD}, and ACG \cite{ACG} can assess robustness by generating falsely predicted adversarial instances. However, they cannot directly evaluate individual fairness because they do not consider similar individuals. Similarly, individual fairness testing approaches, such as ADF \cite{ADF1} and EIDIG \cite{EIDIG}, assess individual fairness by generating adversarial instances with biased predictions. Nevertheless, they do not directly evaluate robustness since they do not take into account the ground truth for the generated instances. Consequently, for the purpose of evaluating false or biased adversarial instances, in the case of robustness evaluation approaches, $N_{f|b}=N_f$, while for individual fairness evaluation approaches, $N_{f|b}=N_b$, as they exclusively assess one aspect of the false or biased adversarial instances. In contrast, RobustFair conducts accurate fairness evaluations of machine learning models from two distinct perspectives: robustness and individual fairness. It generates falsely or biasedly predicted adversarial instances, which can be used not only to assess accurate fairness but also to evaluate robustness and individual fairness simultaneously.

Compared to the performance of state-of-the-art robustness evaluation techniques, RobustFair, on average, identifies a significantly higher number of falsely predicted adversarial instances (3.365-8.183 times) while maintaining a comparable false generation percentage. Similarly, when compared to the performance of state-of-the-art individual fairness evaluation techniques, RobustFair, on average, identifies a significantly higher number of biasedly predicted adversarial instances (2.217-7.818 times). The increased richness of generated false ($N_f$) and biased ($N_b$) adversarial instances by RobustFair can be attributed to the greater diversity of search directions (False Fair, True Bias, and False Bias) compared to other evaluation techniques.  Additionally, RobustFair excels at uncovering issues where robustness and fairness are entangled,  generating numerous false or biased adversarial instances (4.807-15.885 times), which are frequently overlooked in standard robustness assessments and individual fairness evaluations. Moreover, it achieves the highest $R_{f|b}$ generation percentages, ranging from 75.8\% to 95.7\%.


\textit{Answer to Q2: RobustFair employs a combination of False Fair, True Bias, and False Bias perturbations to generate false or biased adversarial instances, enabling a comprehensive evaluation of the accurate fairness of DNNs from multiple perspectives. This multi-perspective perturbation equips RobustFair for accurate fairness evaluation, as well as robustness and individual fairness assessment. With its diverse perturbation directions, RobustFair demonstrates a remarkable capability to generate a significantly larger number of false or biased adversarial instances compared to other evaluation techniques.}

\textit{Q3: How accurate fairness are the baseline models?}

\begin{table}[!htb]
\centering
\caption{Fairness Confusion Matrix Analysis of Techniques across Four Fairness Datasets}
\label{FCM Analysis Statistics}
\begin{tabular}{|l|l|l|l|l|l|l|}
\hline
Technique & \multicolumn{1}{c|}{TF} & \multicolumn{1}{c|}{TB} & \multicolumn{1}{c|}{FF} & \multicolumn{1}{c|}{FB} & \multicolumn{1}{c|}{$sum$} & \multicolumn{1}{c|}{$similarity$} \\ \hline
FGSM & \textbf{68.600} & 14.150 & 5.700 & 4.550 & 93.000 & 0.897 \\ \hline
PGD & 140.400 & 42.850 & 18.350 & 15.150 & 216.750 & 0.897 \\ \hline
APGD & 1946.800 & 1147.700 & 1012.000 & 919.750 & 5026.250 & 0.896 \\ \hline
ACG & 2489.700 & 1122.550 & 568.650 & 617.850 & 4798.750 & 0.898 \\ \hline
ADF & 469.250 & 1642.050 & 20.700 & 400.600 & 2532.600 & 0.901 \\ \hline
EIDIG & 912.350 & 346.550 & 31.400 & 71.500 & 1361.800 & \textbf{0.902} \\ \hline
RobustFair & 1384.700 & \textbf{3979.200} & \textbf{1778.150} & \textbf{1968.400} & \textbf{9110.450} & 0.901 \\ \hline
\end{tabular}
\end{table}

Table~\ref{FCM Analysis Statistics} provides the average statistics of the fairness confusion matrix analysis obtained from 7 trustworthiness evaluation techniques across four fairness datasets (German Credit, Adult, Bank Marketing, and ProPublica Recidivism datasets). To conduct this fairness confusion matrix analysis, we apply fairness data augmentation \cite{AF} to the robustness evaluation results, which involves generating similar counterparts for each adversarial instance. For the individual fairness evaluation results, we approximate the ground truths using ensemble models \cite{EIDIG}. Columns $TF$, $TB$, $FF$, and $FB$ show the numbers of true fair instances, true biased, false fair, and false biased adversarial instances, respectively. Note that by definitions, $FF+FB=N_{f}$ for robustness evaluation, and $TB+FB=N_{b}$ for individual fairness evaluation. The column $similarity$ represents the average of the generated similarity values of these instances. A higher average generated similarity value indicates that the generated instances more closely resemble the original testing set.

It is clear that the RobustFair evaluation approach generates significantly more false fair (on average 6.439 times), false biased (on average 5.820 times), and true biased (on average 5.532 times) adversarial instances compared to other trustworthiness evaluation approaches. Additionally, among the false or biased adversarial instances generated by each technique, the largest number always belongs to true biased adversarial instances, while the number of false biased and false fair adversarial samples is relatively close and both less than the number of true biased adversarial samples. 

Additionally, it is worth noting that the average generated similarity of robustness evaluation approaches (FGSM, PGD, APGD, and ACG) is lower compared to that of individual fairness evaluation approaches (ADF and EIDIG) and the RobustFair approach. This difference can be attributed to the methodology employed by each type of evaluation approach in generating adversarial samples. Robustness evaluation methods perturb all input attributes based on gradients or conjugate gradients for generating adversarial samples. In contrast, individual fairness evaluation methods and RobustFair use the gradients on the instance loss function and its similar counterpart loss function to select specific attributes for perturbation. As a result, individual fairness evaluation methods and RobustFair tend to generate adversarial instances that more closely resemble the test data, resulting in a higher average generated similarity.

\textit{Answer to Q3: The baseline models over the four fairness datasets are most susceptible to true biased perturbations, followed by false biased and false fair perturbations.}

\textit{Q4: How effective is RobustFair for trustworthiness improvement?}

In the next step, we utilize the generated false or biased adversarial instances for trustworthiness improvement. We prioritize adversarial instances generated from the training dataset that closely resemble the original training set for training dataset augmentation. This process involves selecting instances whose generative similarity $S_{(v_p,y_p)}$ exceeds a predefined threshold $\sigma$. Then the augmented dataset is utilized to retrain the original DNNs to achieve higher levels of accurate fairness.

\begin{figure}[!htb]
    \begin{minipage}[t]{0.24\linewidth}
        \centering
        \includegraphics[width=1.0\linewidth]{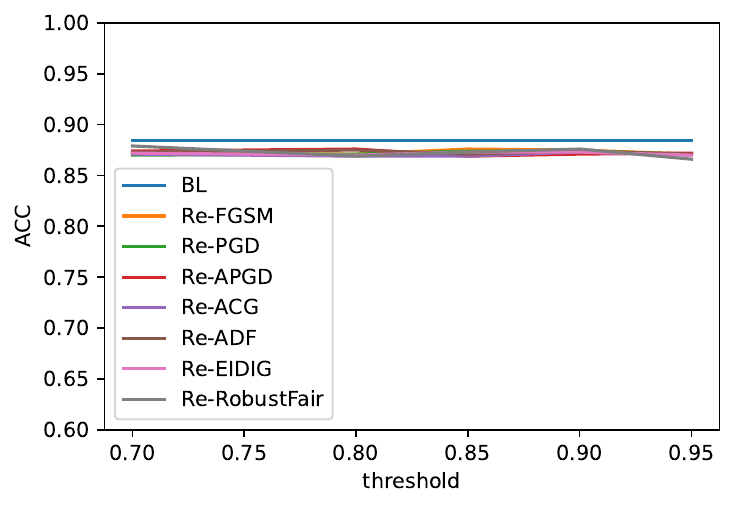}
    \end{minipage}
    \begin{minipage}[t]{0.24\linewidth}
        \centering
        \includegraphics[width=1.0\linewidth]{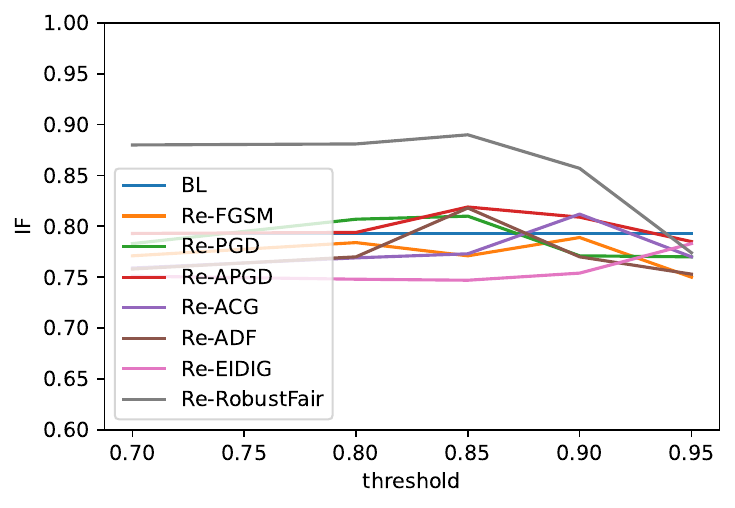}
    \end{minipage}
    \begin{minipage}[t]{0.24\linewidth}
        \centering
        \includegraphics[width=1.0\linewidth]{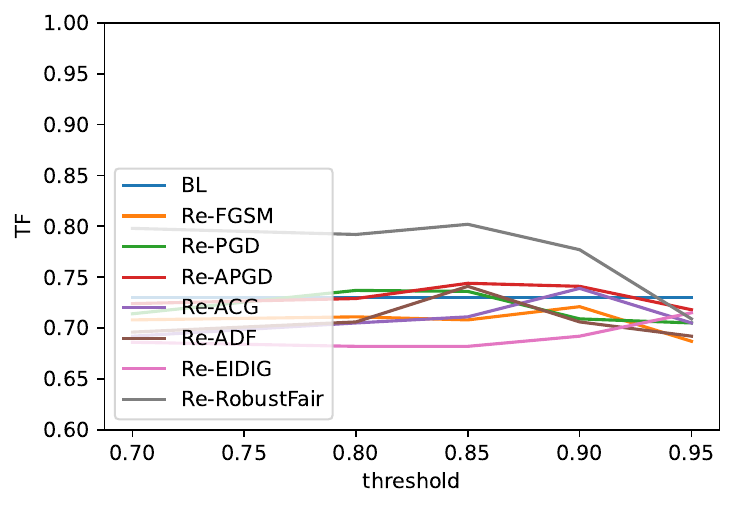}
    \end{minipage}
    \begin{minipage}[t]{0.24\linewidth}
        \centering
        \includegraphics[width=1.0\linewidth]{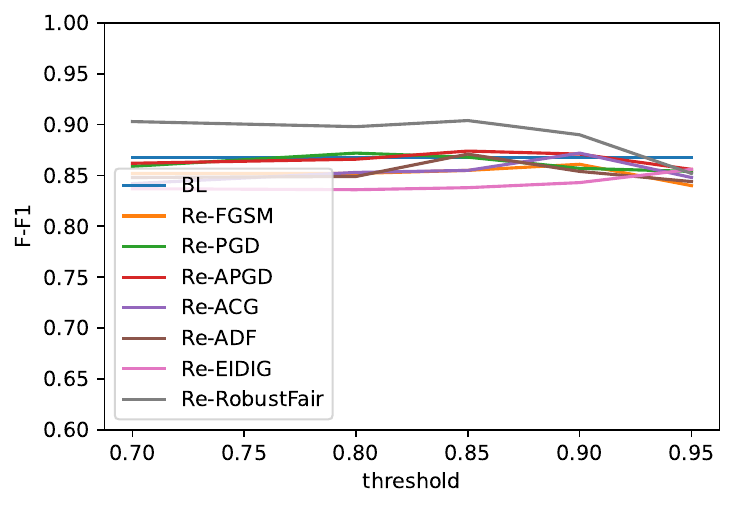}
    \end{minipage}

    \caption{Accuracy (ACC), Individual Fairness (IF), and Accurate Fairness (TF, F-F1) of Retrained Models}
    \label{retrained ACC, IF, AF}
\end{figure}

\figureautorefname~\ref{retrained ACC, IF, AF} presents the average performance of the baseline models (BL) and the retrained models (Re-$M$), across four datasets. We consider various thresholds for the selection of adversarial instances ($\sigma \in \{0.70, 0.75, 0.80, 0.85, 0.9, 0.95\}$). In this context, Re-$M$ denotes a model retrained using the adversarial instances detected by technique $M$ from the training dataset. The evaluation metrics used include Accuracy (ACC), Individual Fairness (IF), True Fairness (TF), and Fair-F1 Score (F-F1), which quantifies the relationship between accuracy and individual fairness \cite{AF}. Higher values of ACC, IF, TF and F-F1 indicate superior performance in terms of accuracy, individual fairness, and accurate fairness, respectively. Notably, during the process of training dataset augmentation, RobustFair and the state-of-the-art evaluation techniques estimate the ground truths of the generated instances in different ways: FGSM, PGD, APGD, and ACG reuse the original ground truths according to classical robustness requirements; ADF and EIDIG determine the ground truths using ensemble models \cite{EIDIG}. In contrast, RobustFair employs the total derivative for ground truth approximation.

An interesting observation in \figureautorefname~\ref{retrained ACC, IF, AF} is that the individual fairness and accuracy fairness of retrained models initially increase and then decrease as $\sigma$ increases. This phenomenon can be explained as the impact of the selection threshold $\sigma$. As $\sigma$ increases, more adversarial instances are incorporated into the retraining process, providing more adversarial information on individual fairness and accurate fairness to improve trustworthiness. However, higher $\sigma$ values also allow adversarial instances with less similarity to the original training set to be included, introducing noise into the retraining process. It is noteworthy that most retrained models achieve their best trustworthiness improvement when $\sigma$ is set to 0.85.

\begin{table}[!htb]
\centering
\caption{Statistics of Retrained Models across Four Fairness Datasets, $\sigma=0.85$}
\label{retrain model statistics}
\begin{tabular}{|l|c|c|c|c|}
\hline
\multicolumn{1}{|c|}{Model} & Acc & IF & TF & F-F1 \\ \hline
BL & \textbf{88.4\%} & 79.3\% & 73.0\% & 0.868 \\ \hline
Re-FGSM & 87.6\% & 77.1\% & 70.8\% & 0.855 \\ \hline
Re-PGD & 87.4\% & 81.0\% & 73.6\% & 0.868 \\ \hline
Re-APGD & 86.9\% & 81.9\% & 74.4\% & 0.874 \\ \hline
Re-ACG & 86.9\% & 77.3\% & 71.1\% & 0.855 \\ \hline
Re-ADF & 87.1\% & 81.8\% & 74.1\% & 0.871 \\ \hline
Re-EIDIG & 87.3\% & 74.7\% & 68.2\% & 0.838 \\ \hline
Re-RobustFair & 87.3\% & \textbf{89.0\%} & \textbf{80.2\%} & \textbf{0.904} \\ \hline
\end{tabular}
\end{table}

Table~\ref{retrain model statistics} presents the statistics obtained from the baseline models and retrained models across four fairness datasets, with an adversarial instance selection threshold of $\sigma=0.85$. The Re-RobustFair models exhibit the most substantial improvements in accurate fairness, with an average increase of 7.2\% in TF and 0.036 in F-F1. Additionally, they also achieve a remarkable 9.7\% average improvement in individual fairness without compromising accuracy across all four datasets. This highlights the crucial role of RobustFair in enhancing trustworthiness by effectively promoting accurate fairness and individual fairness while maintaining accuracy.

\textit{Answer to Q4: The adversarial instances identified by RobustFair outperform those identified by the other testing approaches in trustworthiness improvement. By carefully choosing an appropriate selection threshold $\sigma$, the retrained model can achieve maximum improvement in trustworthiness, in terms of both accurate fairness and individual fairness, while maintaining accuracy.}

\textit{Q5: Global Generation VS local Generation}

Finally, we conducted an analysis of global and local generations. Table~\ref{Global and Local Comparison} presents a comparative evaluation of performance across four datasets for both global and local generation. To facilitate this comparison, we applied K-Means clustering to the test dataset and selected 100 initial seeds from the resulting clusters for ADF, EIDIG, and RobustFair techniques. It's worth noting that robustness evaluation approaches do not involve global and local search, making them unsuitable for this experiment. We carried out two sets of experiments: one with 10 global iterations and 0 local iterations, and another with 0 global iterations and 10 local iterations, to obtain the results of global or local generation using identical inputs. Given that fairness evaluation methods may not comprehensively identify false or biased adversarial instances, our focus is on comparing the counts and percentages of biased adversarial instances ($N_b$ and $R_b$) in the generated results.


\begin{table}[!htb]
\centering
\caption{Global and Local Generation Comparison across Four Fairness Datasets}
\label{Global and Local Comparison}
\begin{tabular}{|l|lll|lll|}
\hline
\multirow{2}{*}{Techniques} & \multicolumn{3}{c|}{Global Generation} & \multicolumn{3}{c|}{Local Generation} \\ \cline{2-7} 
 & \multicolumn{1}{c|}{$R_b$} & \multicolumn{1}{c|}{$N_b$} & $sum$ & \multicolumn{1}{c|}{$R_b$} & \multicolumn{1}{c|}{$N_b$} & $sum$ \\ \hline
ADF & \multicolumn{1}{c|}{\textbf{50.2\%}} & \multicolumn{1}{l|}{247.050} & 492.850 & \multicolumn{1}{c|}{62.2\%} & \multicolumn{1}{l|}{165.000} & 262.200 \\ \hline
EIDIG & \multicolumn{1}{c|}{30.7\%} & \multicolumn{1}{l|}{48.050} & 160.650 & \multicolumn{1}{c|}{22.6\%} & \multicolumn{1}{l|}{190.550} & \textbf{842.450} \\ \hline
RobustFair & \multicolumn{1}{c|}{39.1\%} & \multicolumn{1}{l|}{\textbf{570.550}} & \textbf{1644.150} & \multicolumn{1}{c|}{\textbf{74.9\%}} & \multicolumn{1}{l|}{\textbf{227.650}} & 300.150 \\ \hline
\end{tabular}
\end{table}

As observed, global generation tends to search for significantly more times compared to local search. This difference can be attributed to the distinct objectives of these two approaches. The global search focuses on generating a variety of biased instances by interfering with as many non-sensitive features as possible. In contrast, local search aims to make minimal changes to the model's input to preserve the original discrimination in the predictions. This is quantitatively achieved by changing the non-sensitive features as little as possible. It is also worth noting that our method generates the most biased instances in both global (3.867 times) and local (1.281 times) generations. This is because our method searches for biased samples from two search directions: true biased and false biased directions.

\textit{Answer to Q5: The global search aims to generate diverse biased instances, while the local search intends to minimally change the model outputs to maintain the discrimination in original predictions.}

\section{Conclusion and Future Work}

In this paper, we introduce RobustFair, an innovative approach for evaluating the accurate fairness of DNNs in the context of intertwined issues of robustness and individual fairness. RobustFair leverages the fairness confusion matrix induced by accurate fairness to guide distinct gradient search directions for generating false fair, true biased, and false biased adversarial perturbations. These perturbations have a dual impact on adversarial instances and their similar counterparts to either undermine prediction accuracy (robustness) or cause biased predictions (individual fairness). RobustFair goes further by approximating loss function values for adversarial instances and inferring their ground truth based on these approximations. These generated adversarial instances can then be utilized for data augmentation and model retraining, enhancing the trustworthiness of the original DNNs.

We conduct comparative case studies using benchmark fairness datasets, demonstrating the synergistic effectiveness of RobustFair in accurate fairness evaluation, particularly in detecting subtle robustness defects entangled with individual fairness concerns. RobustFair surpasses state-of-the-art approaches in both the quantity and quality of diverse adversarial instances detected. Furthermore, the adversarial instances identified by RobustFair outperform those identified by the other testing approaches in trustworthiness improvement, enhancing the accurate fairness and individual fairness of DNNs without compromising overall accuracy.

While RobustFair is a white-box testing approach that requires access to the gradients of DNNs, an interesting avenue for future work is to extend accurate fairness evaluation to black-box scenarios. Additionally, RobustFair can potentially be expanded to integrate with other trustworthiness evaluation approaches, accommodating various datasets, models, and tasks in practical applications.

\newpage

\bibliographystyle{unsrt}  
\bibliography{templateArxiv}  

\begin{thebibliography}{10}

\bibitem{Adv_E2}
Ian~J. Goodfellow, Jonathon Shlens, and Christian Szegedy.
\newblock Explaining and harnessing adversarial examples.
\newblock In Yoshua Bengio and Yann LeCun, editors, {\em 3rd International
  Conference on Learning Representations, {ICLR} 2015, San Diego, CA, USA, May
  7-9, 2015, Conference Track Proceedings}, 2015.

\bibitem{IF1}
Cynthia Dwork, Moritz Hardt, Toniann Pitassi, Omer Reingold, and Richard~S.
  Zemel.
\newblock Fairness through awareness.
\newblock In Shafi Goldwasser, editor, {\em Innovations in Theoretical Computer
  Science 2012, Cambridge, MA, USA, January 8-10, 2012}, pages 214--226. {ACM},
  2012.

\bibitem{OOD1}
Zhen Zhang, Peng Wu, Yuhang Chen, and Jing Su.
\newblock Out-of-distribution detection through relative
  activation-deactivation abstractions.
\newblock In Zhi Jin, Xuandong Li, Jianwen Xiang, Leonardo Mariani, Ting Liu,
  Xiao Yu, and Nahgmeh Ivaki, editors, {\em 32nd {IEEE} International Symposium
  on Software Reliability Engineering, {ISSRE} 2021, Wuhan, China, October
  25-28, 2021}, pages 150--161. {IEEE}, 2021.

\bibitem{R_IF1}
Alex Gittens, B{\"{u}}lent Yener, and Moti Yung.
\newblock An adversarial perspective on accuracy, robustness, fairness, and
  privacy: Multilateral-tradeoffs in trustworthy {ML}.
\newblock {\em {IEEE} Access}, 10:120850--120865, 2022.

\bibitem{R_IF2}
Jae{-}Gil Lee, Yuji Roh, Hwanjun Song, and Steven~Euijong Whang.
\newblock Machine learning robustness, fairness, and their convergence.
\newblock In Feida Zhu, Beng~Chin Ooi, and Chunyan Miao, editors, {\em {KDD}
  '21: The 27th {ACM} {SIGKDD} Conference on Knowledge Discovery and Data
  Mining, Virtual Event, Singapore, August 14-18, 2021}, pages 4046--4047.
  {ACM}, 2021.

\bibitem{Adv2}
Jing Su, Zhen Zhang, Peng Wu, Xuran Li, and Jian Zhang.
\newblock Adversarial input detection based on critical transformation
  robustness.
\newblock In {\em {IEEE} 33rd International Symposium on Software Reliability
  Engineering, {ISSRE} 2022, Charlotte, NC, USA, October 31 - Nov. 3, 2022},
  pages 390--401. {IEEE}, 2022.

\bibitem{ADF}
Peixin Zhang, Jingyi Wang, Jun Sun, Guoliang Dong, Xinyu Wang, Xingen Wang,
  Jin~Song Dong, and Ting Dai.
\newblock White-box fairness testing through adversarial sampling.
\newblock In Gregg Rothermel and Doo{-}Hwan Bae, editors, {\em {ICSE} '20: 42nd
  International Conference on Software Engineering, Seoul, South Korea, 27 June
  - 19 July, 2020}, pages 949--960. {ACM}, 2020.

\bibitem{banking}
S.~Moro, P.~Rita, and P.~Cortez.
\newblock {Bank Marketing}.
\newblock UCI Machine Learning Repository, 2012.

\bibitem{adult}
{Adult}.
\newblock UCI Machine Learning Repository, 1996.
\newblock {DOI}: \url{10.24432/C5XW20}.

\bibitem{COMPAS}
Julia Angwin, Jeff Larson, Surya Mattu, and Lauren Kirchner.
\newblock Machine bias.
\newblock {\em Ethics of Data and Analytics: Concepts and Cases}, page 254,
  2022.

\bibitem{AF}
Xuran Li, Peng Wu, and Jing Su.
\newblock Accurate fairness: Improving individual fairness without trading
  accuracy.
\newblock In Brian Williams, Yiling Chen, and Jennifer Neville, editors, {\em
  Thirty-Seventh {AAAI} Conference on Artificial Intelligence, {AAAI} 2023,
  Thirty-Fifth Conference on Innovative Applications of Artificial
  Intelligence, {IAAI} 2023, Thirteenth Symposium on Educational Advances in
  Artificial Intelligence, {EAAI} 2023, Washington, DC, USA, February 7-14,
  2023}, pages 14312--14320. {AAAI} Press, 2023.

\bibitem{credit}
Hans Hofmann.
\newblock {Statlog (German Credit Data)}.
\newblock UCI Machine Learning Repository, 1994.
\newblock {DOI}: \url{10.24432/C5NC77}.

\bibitem{activation}
Vinod Nair and Geoffrey~E. Hinton.
\newblock Rectified linear units improve restricted boltzmann machines.
\newblock In Johannes F{\"{u}}rnkranz and Thorsten Joachims, editors, {\em
  Proceedings of the 27th International Conference on Machine Learning
  (ICML-10), June 21-24, 2010, Haifa, Israel}, pages 807--814. Omnipress, 2010.

\bibitem{FGSM}
Ian~J. Goodfellow, Jonathon Shlens, and Christian Szegedy.
\newblock Explaining and harnessing adversarial examples.
\newblock In Yoshua Bengio and Yann LeCun, editors, {\em 3rd International
  Conference on Learning Representations, {ICLR} 2015, San Diego, CA, USA, May
  7-9, 2015, Conference Track Proceedings}, 2015.

\bibitem{R3}
Muhammad Awais and Sung{-}Ho Bae.
\newblock A survey on efficient methods for adversarial robustness.
\newblock {\em {IEEE} Access}, 10:118815--118830, 2022.

\bibitem{PGD}
Aleksander Madry, Aleksandar Makelov, Ludwig Schmidt, Dimitris Tsipras, and
  Adrian Vladu.
\newblock Towards deep learning models resistant to adversarial attacks.
\newblock In {\em 6th International Conference on Learning Representations,
  {ICLR} 2018, Vancouver, BC, Canada, April 30 - May 3, 2018, Conference Track
  Proceedings}. OpenReview.net, 2018.

\bibitem{APGD}
Francesco Croce and Matthias Hein.
\newblock Reliable evaluation of adversarial robustness with an ensemble of
  diverse parameter-free attacks.
\newblock In {\em Proceedings of the 37th International Conference on Machine
  Learning, {ICML} 2020, 13-18 July 2020, Virtual Event}, volume 119 of {\em
  Proceedings of Machine Learning Research}, pages 2206--2216. {PMLR}, 2020.

\bibitem{ACG}
Keiichiro Yamamura, Haruki Sato, Nariaki Tateiwa, Nozomi Hata, Toru Mitsutake,
  Issa Oe, Hiroki Ishikura, and Katsuki Fujisawa.
\newblock Diversified adversarial attacks based on conjugate gradient method.
\newblock In Kamalika Chaudhuri, Stefanie Jegelka, Le~Song, Csaba
  Szepesv{\'{a}}ri, Gang Niu, and Sivan Sabato, editors, {\em International
  Conference on Machine Learning, {ICML} 2022, 17-23 July 2022, Baltimore,
  Maryland, {USA}}, volume 162 of {\em Proceedings of Machine Learning
  Research}, pages 24872--24894. {PMLR}, 2022.

\bibitem{GTest1}
Nian Si, Karthyek Murthy, Jose~H. Blanchet, and Viet~Anh Nguyen.
\newblock Testing group fairness via optimal transport projections.
\newblock In Marina Meila and Tong Zhang, editors, {\em Proceedings of the 38th
  International Conference on Machine Learning, {ICML} 2021, 18-24 July 2021,
  Virtual Event}, volume 139 of {\em Proceedings of Machine Learning Research},
  pages 9649--9659. {PMLR}, 2021.

\bibitem{GTest2}
Furkan Gursoy and Ioannis~A. Kakadiaris.
\newblock Error parity fairness: Testing for group fairness in regression
  tasks.
\newblock {\em CoRR}, abs/2208.08279, 2022.

\bibitem{ADF1}
Peixin Zhang, Jingyi Wang, Jun Sun, Xinyu Wang, Guoliang Dong, Xingen Wang,
  Ting Dai, and Jin~Song Dong.
\newblock Automatic fairness testing of neural classifiers through adversarial
  sampling.
\newblock {\em {IEEE} Trans. Software Eng.}, 48(9):3593--3612, 2022.

\bibitem{EIDIG}
Lingfeng Zhang, Yueling Zhang, and Min Zhang.
\newblock Efficient white-box fairness testing through gradient search.
\newblock In Cristian Cadar and Xiangyu Zhang, editors, {\em {ISSTA} '21: 30th
  {ACM} {SIGSOFT} International Symposium on Software Testing and Analysis,
  Virtual Event, Denmark, July 11-17, 2021}, pages 103--114. {ACM}, 2021.

\bibitem{Mathematical_Analysis}
John~E Hutchinson and Richard~J Loy.
\newblock Introduction to mathematical analysis.
\newblock {\em School of Mathematical Sciences}, 1995.

\bibitem{quantile_regression}
Roger Koenker and Kevin~F Hallock.
\newblock Quantile regression.
\newblock {\em Journal of economic perspectives}, 15(4):143--156, 2001.

\bibitem{KMeans}
Stuart~P. Lloyd.
\newblock Least squares quantization in {PCM}.
\newblock {\em {IEEE} Trans. Inf. Theory}, 28(2):129--136, 1982.

\end{thebibliography}

\clearpage
\appendix
\section{More Experimental Results}
\subsection{Loss Function Trends During Fairness Confusion Perturbations}

\begin{figure}[!htb]
\centering
    \begin{minipage}[t]{0.25\linewidth}
        \centering
        \includegraphics[width=1.0\linewidth]{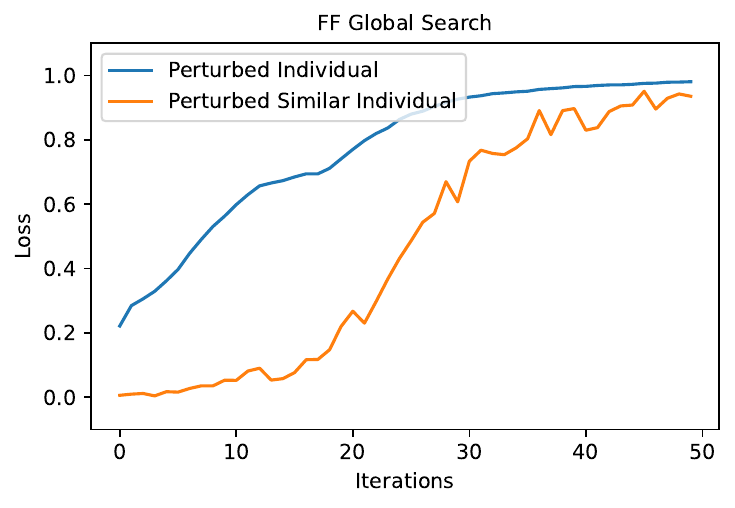}
    \end{minipage}
    \begin{minipage}[t]{0.25\linewidth}
        \centering
        \includegraphics[width=1.0\linewidth]{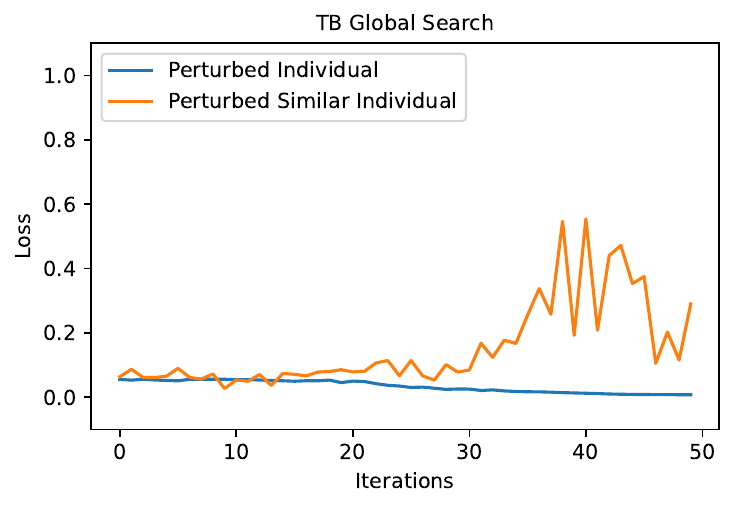}
    \end{minipage}
    \begin{minipage}[t]{0.25\linewidth}
        \centering
        \includegraphics[width=1.0\linewidth]{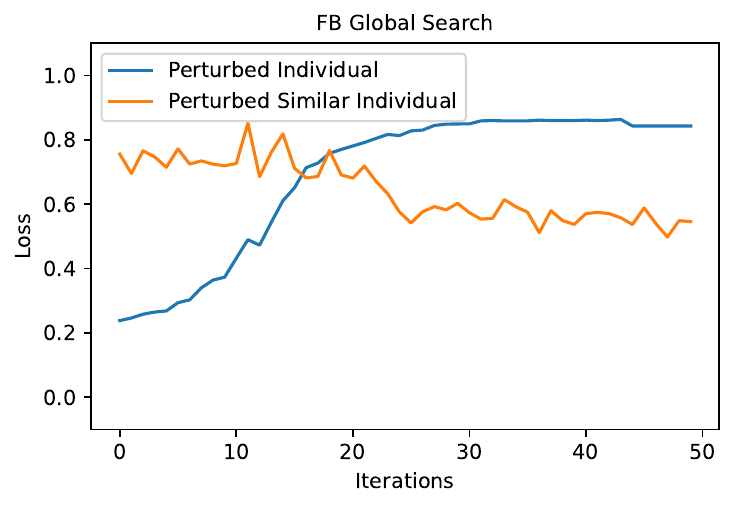}
    \end{minipage}
    \caption{Global Search Process on Adult Dataset}
    \label{Adult G search}

    \begin{minipage}[t]{0.25\linewidth}
        \centering
        \includegraphics[width=1.0\linewidth]{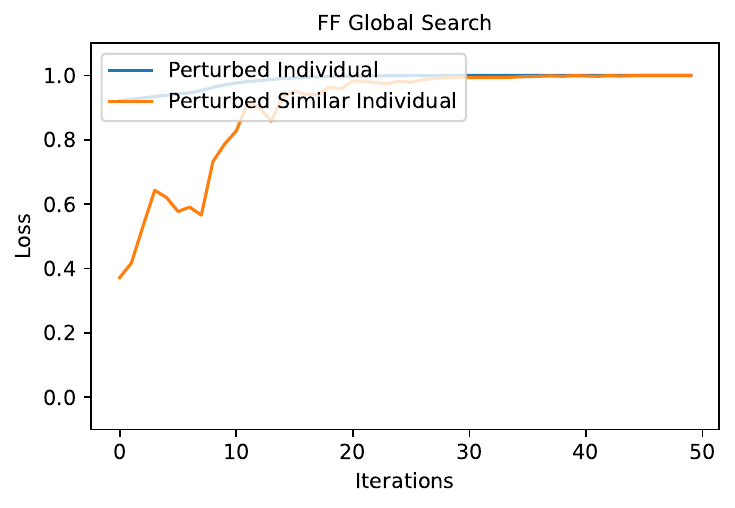}
    \end{minipage}
    \begin{minipage}[t]{0.25\linewidth}
        \centering
        \includegraphics[width=1.0\linewidth]{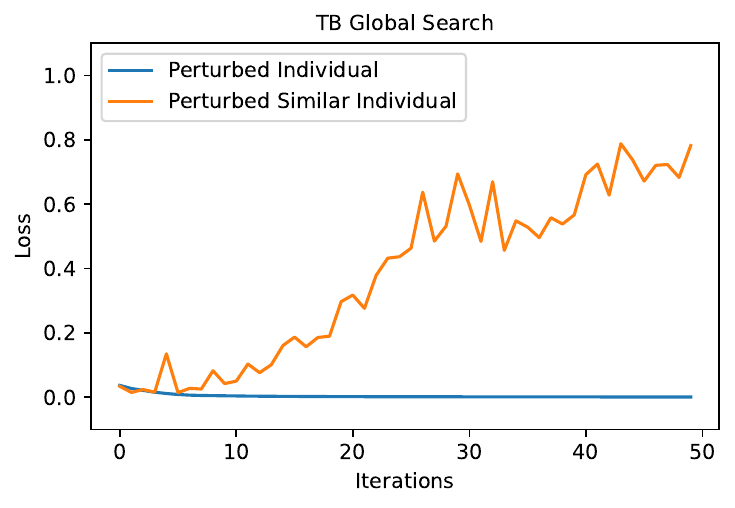}
    \end{minipage}
    \begin{minipage}[t]{0.25\linewidth}
        \centering
        \includegraphics[width=1.0\linewidth]{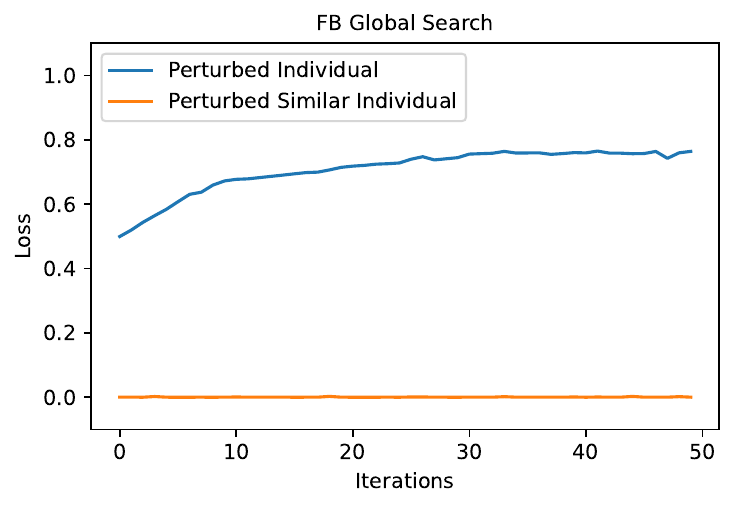}
    \end{minipage}
    \caption{Global Search Process on COMPAS Dataset}
    \label{compas G search}

    \begin{minipage}[t]{0.25\linewidth}
        \centering
        \includegraphics[width=1.0\linewidth]{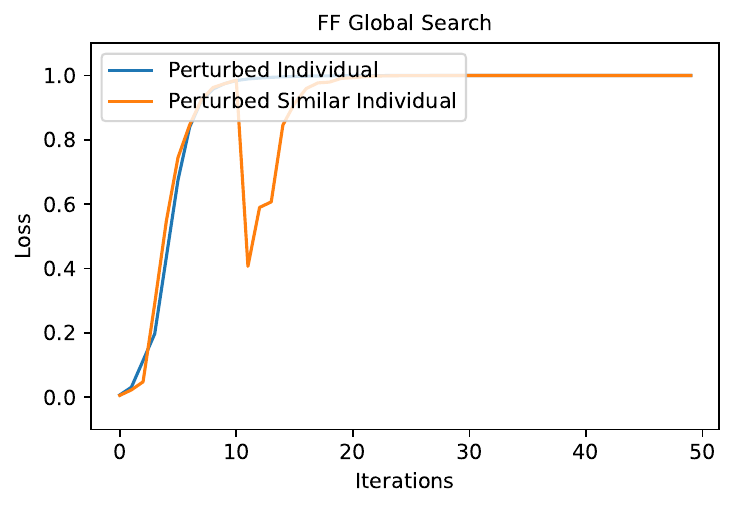}
    \end{minipage}
    \begin{minipage}[t]{0.25\linewidth}
        \centering
        \includegraphics[width=1.0\linewidth]{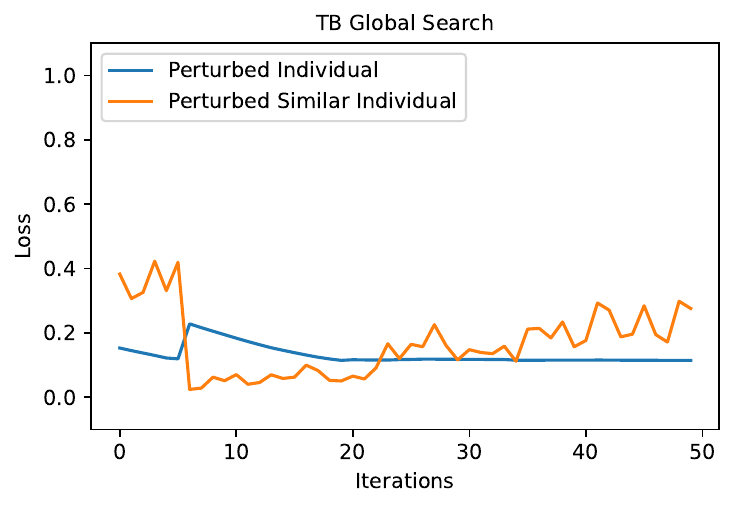}
    \end{minipage}
    \begin{minipage}[t]{0.25\linewidth}
        \centering
        \includegraphics[width=1.0\linewidth]{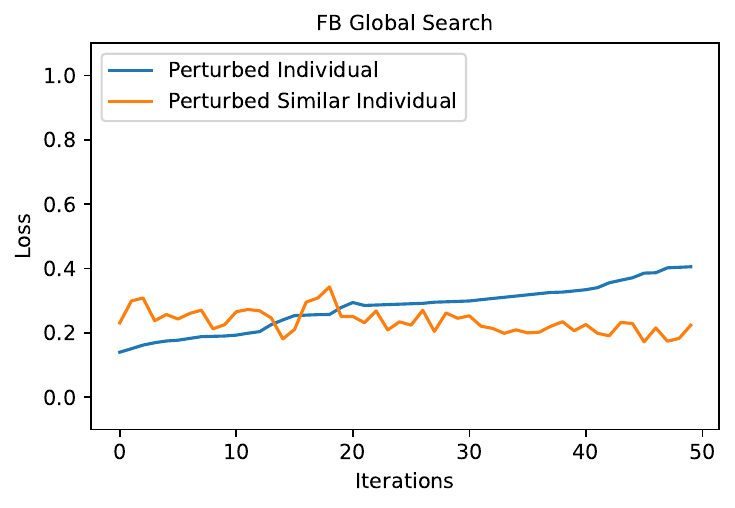}
    \end{minipage}
    \caption{Global Search Process on German Credit Dataset}
    \label{credit G search}
\end{figure}

Figures \ref{Adult G search}, \ref{compas G search}, and \ref{credit G search} depict the trends in the loss function for an instance and its corresponding counterpart during the global generation of fairness confusion perturbations (false fair, true bias, and false bias). These figures were generated by tracking the loss function values of one instance from the Adult, COMPAS, or German Credit datasets, respectively, when subjected to perturbations. The perturbation parameters used were a step size of $p=0.01$ and 50 iterations.

As shown in Figures \ref{Adult G search}, \ref{compas G search}, and \ref{credit G search}, we observe similar trends across the three datasets during the global generation process. False fair perturbations have a similar impact on both the instance and its counterpart, resulting in an increase in their loss functions. This suggests that the perturbed instance may receive a false treatment according to its ground truth while remaining similar to its counterpart. In contrast, during true biased and false biased global generation, adversarial perturbations affect the instance and its counterpart differently. One experiences an increase in loss, while the other undergoes a decrease, indicating differential treatment of the perturbed instance compared to its counterpart.

\begin{figure}[!htb]
\centering
    \begin{minipage}[t]{0.25\linewidth}
        \centering
        \includegraphics[width=1.0\linewidth]{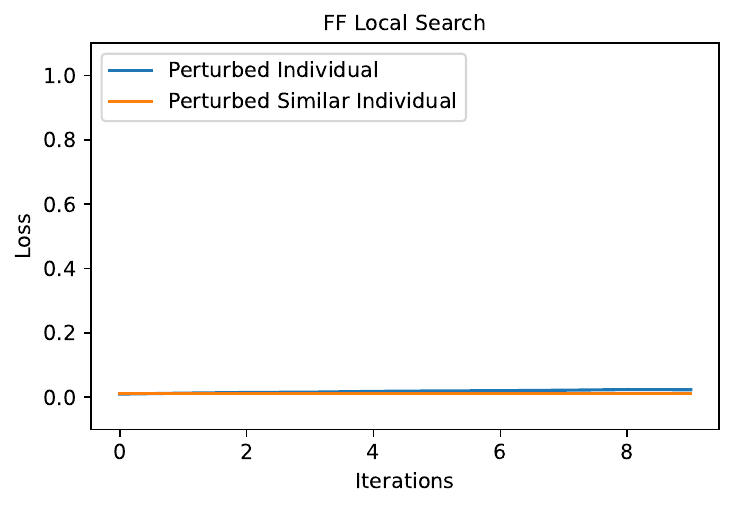}
    \end{minipage}
    \begin{minipage}[t]{0.25\linewidth}
        \centering
        \includegraphics[width=1.0\linewidth]{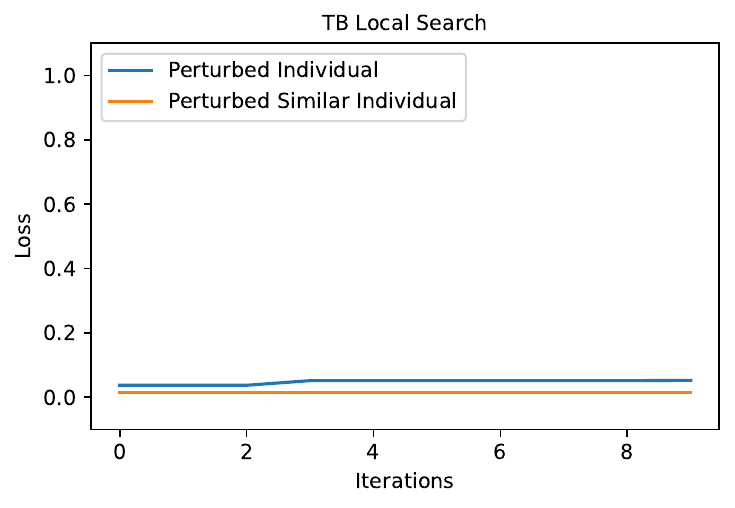}
    \end{minipage}
    \begin{minipage}[t]{0.25\linewidth}
        \centering
        \includegraphics[width=1.0\linewidth]{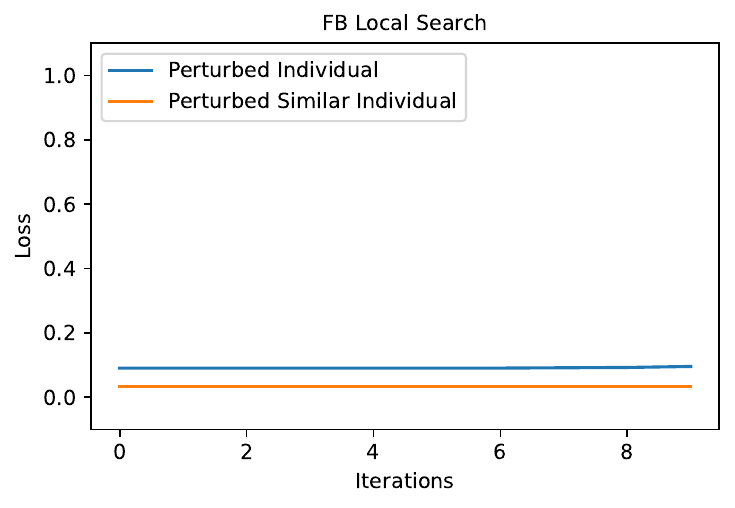}
    \end{minipage}
    \caption{Local Search Process on Adult Dataset}
    \label{Adult L search}

    \begin{minipage}[t]{0.25\linewidth}
        \centering
        \includegraphics[width=1.0\linewidth]{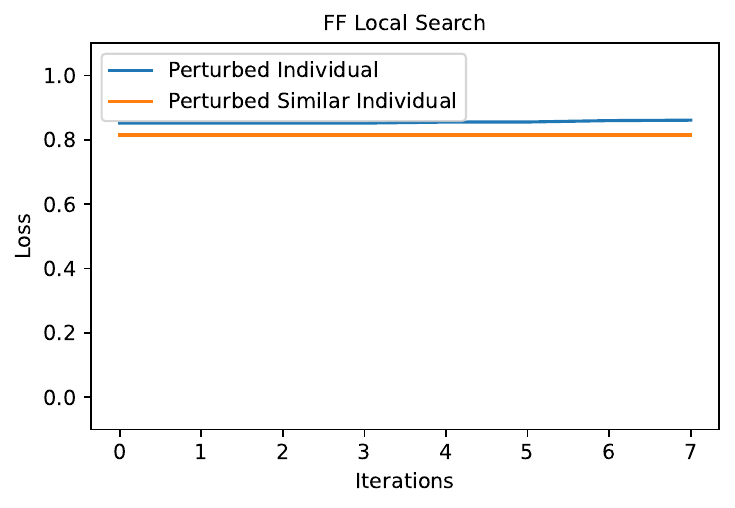}
    \end{minipage}
    \begin{minipage}[t]{0.25\linewidth}
        \centering
        \includegraphics[width=1.0\linewidth]{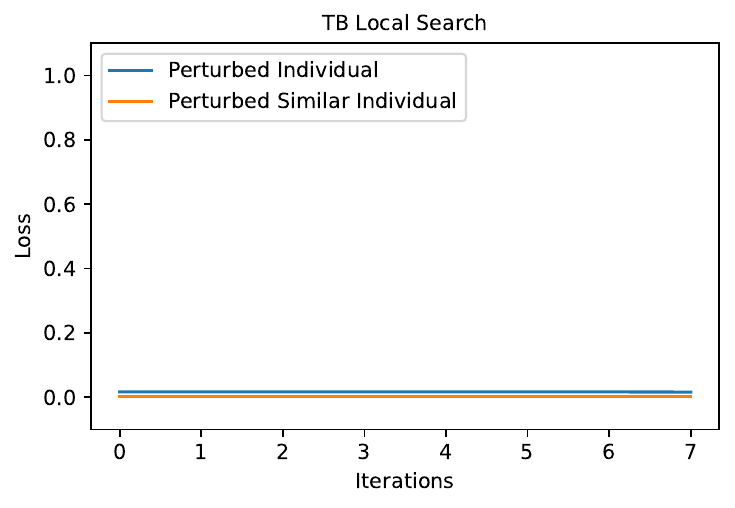}
    \end{minipage}
    \begin{minipage}[t]{0.25\linewidth}
        \centering
        \includegraphics[width=1.0\linewidth]{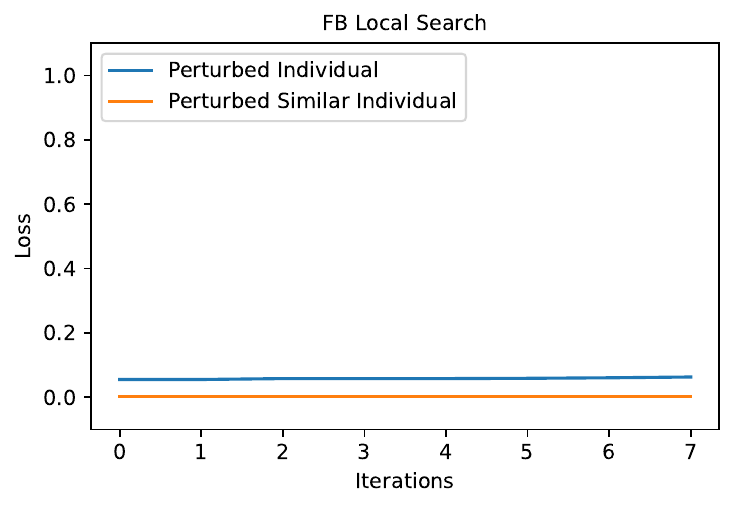}
    \end{minipage}
    \caption{Local Search Process on COMPAS Dataset}
    \label{compas L search}

    \begin{minipage}[t]{0.25\linewidth}
        \centering
        \includegraphics[width=1.0\linewidth]{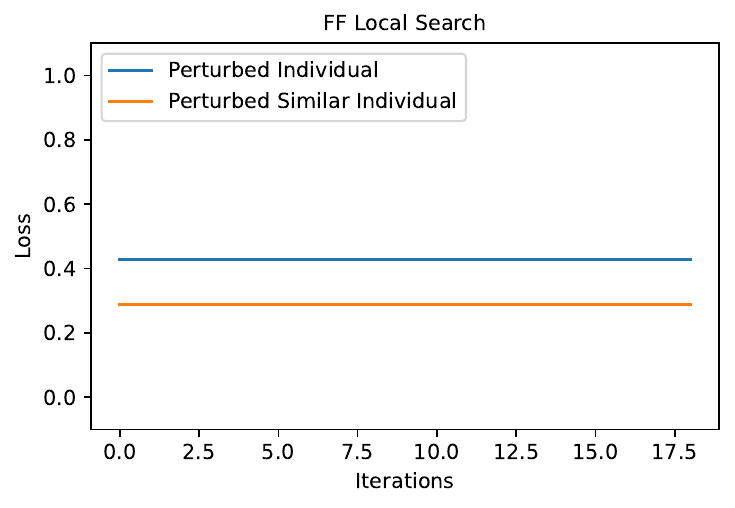}
    \end{minipage}
    \begin{minipage}[t]{0.25\linewidth}
        \centering
        \includegraphics[width=1.0\linewidth]{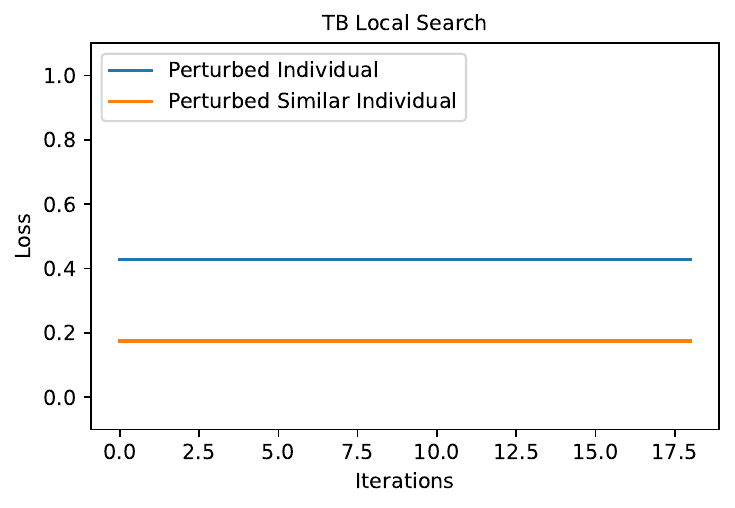}
    \end{minipage}
    \begin{minipage}[t]{0.25\linewidth}
        \centering
        \includegraphics[width=1.0\linewidth]{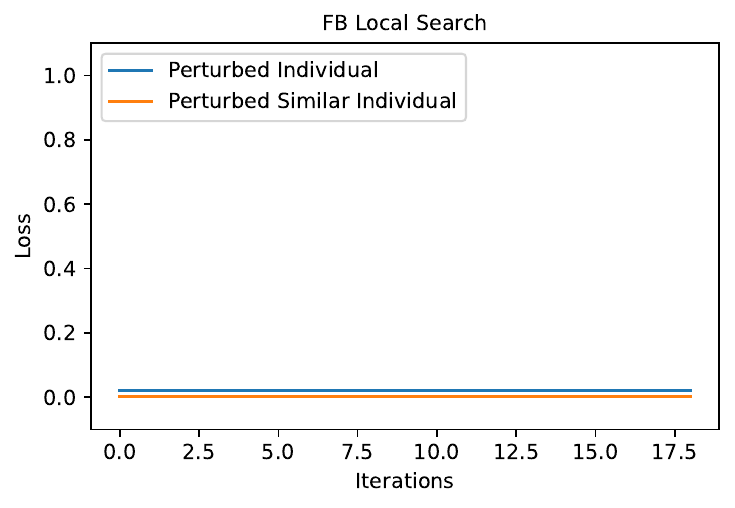}
    \end{minipage}
    \caption{Local Search Process on German Credit Dataset}
    \label{credit L search}
\end{figure}

Figures \ref{Adult L search}, \ref{compas L search}, and \ref{credit L search} depict the changes in the loss function during the local generation process. It's evident that local perturbations have minimal effects on both the instance and its similar counterpart. This occurs because the local process aims to identify false or biased adversarial instances in proximity to the input instance. During this process, we arrange the non-sensitive attributes in descending order based on their absolute partial derivative values and sequentially perturb these attributes. These perturbations follow the guidance of the fairness confusion matrix, with the objective of minimizing alterations to the accurate fairness of the predictions.

\subsection{Quantile Regression Analysis of Loss Function Trends}

In addition, we employ quantile regression (Quantile Regression \cite{quantile_regression}) to analyze the trends in the loss function of 100 instances from the Adult, COMPAS, and German Credit datasets. Initially, we apply K-Means clustering (K-Means \cite{KMeans}) to the test dataset and meticulously select 100 initial seeds from the resulting clusters. Subsequently, we record the loss function values of these instances when subjected to the global generation and local generation processes of fairness confusion perturbations.

Regarding the loss function as a random variable, given the number of iterations, we employ quantile regression to model the median value of the loss function, denoted as $l$, as a linear function of the number of iterations, represented as: $\mathbb{Q} \frac{1}{2}\left( l|iter \right)= \beta_0 +\beta_1\cdot iter$. Here, $\mathbb{Q}{\frac{1}{2}}(l | \text{iter})$ represents the median of the loss function given the number of iterations, and $\beta=[\beta_0, \beta_1]^T$ is the parameter vector estimated through the optimization problem expressed as: $\hat{\beta}=\underset{\beta \in R^2}{\mathrm{arg}\min}\sum_{k=1}^n{\left| l_k-\beta_0-\beta_1 \cdot iter_k \right|}$.

\begin{figure}[!htb]
\centering
    \begin{minipage}[t]{0.25\linewidth}
        \centering
        \includegraphics[width=1.0\linewidth]{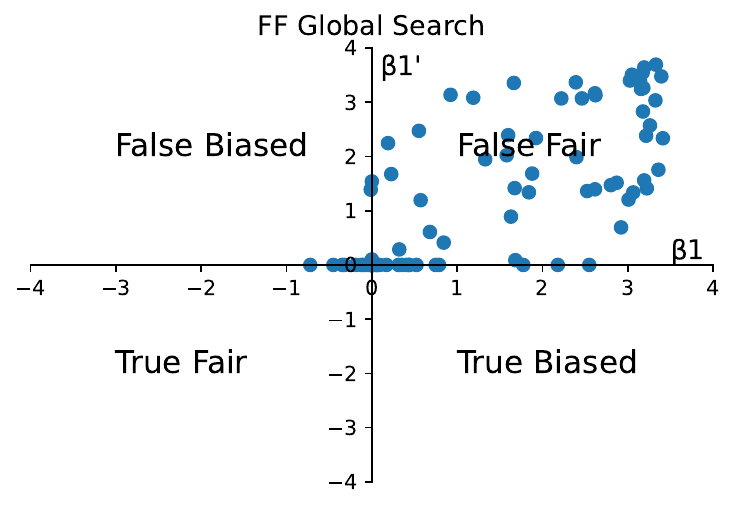}
    \end{minipage}
    \begin{minipage}[t]{0.25\linewidth}
        \centering
        \includegraphics[width=1.0\linewidth]{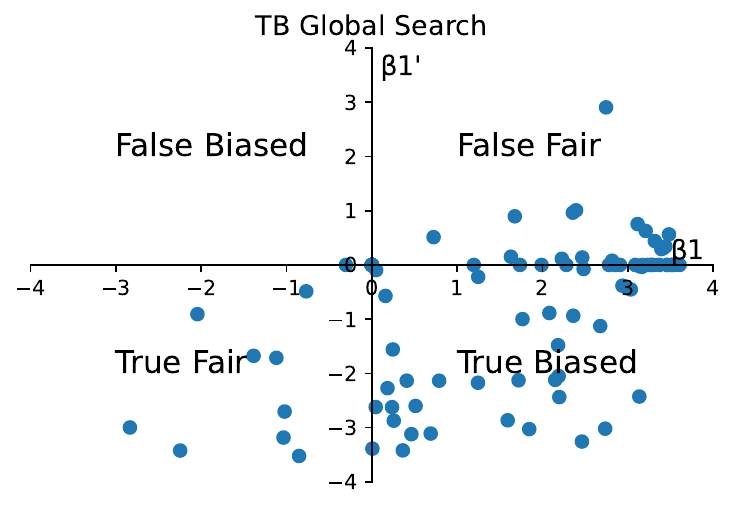}
    \end{minipage}
    \begin{minipage}[t]{0.25\linewidth}
        \centering
        \includegraphics[width=1.0\linewidth]{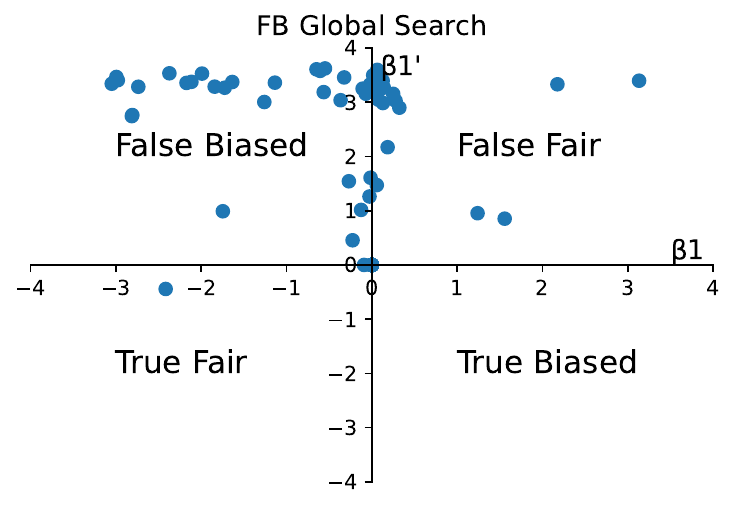}
    \end{minipage}

    \caption{Global Fairness Confusion Perturbation Analysis on Adult Dataset}
    \label{adult global analysis}
\end{figure}

\begin{figure}[!htb]
\centering
    \begin{minipage}[t]{0.25\linewidth}
        \centering
        \includegraphics[width=1.0\linewidth]{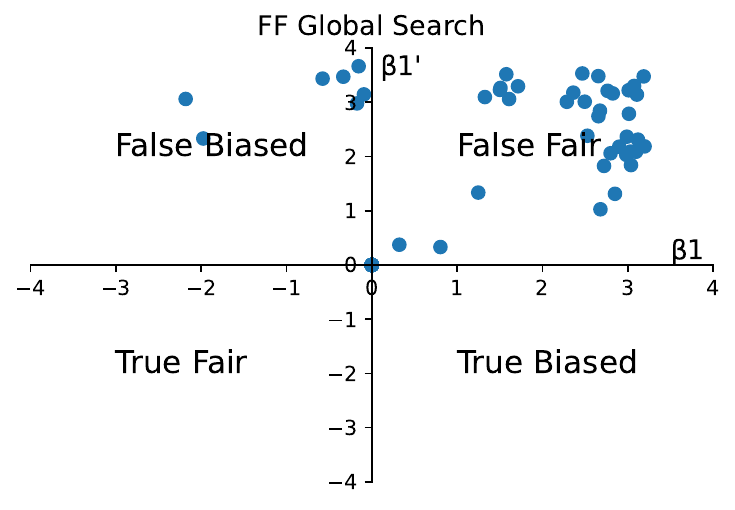}
    \end{minipage}
    \begin{minipage}[t]{0.25\linewidth}
        \centering
        \includegraphics[width=1.0\linewidth]{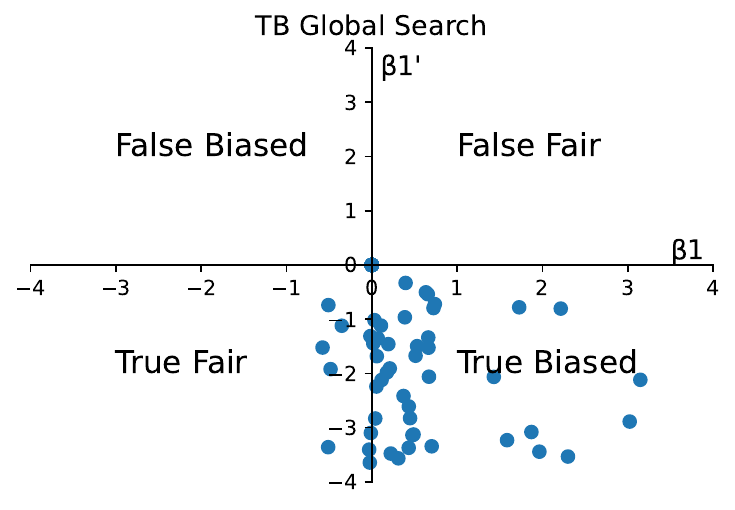}
    \end{minipage}
    \begin{minipage}[t]{0.25\linewidth}
        \centering
        \includegraphics[width=1.0\linewidth]{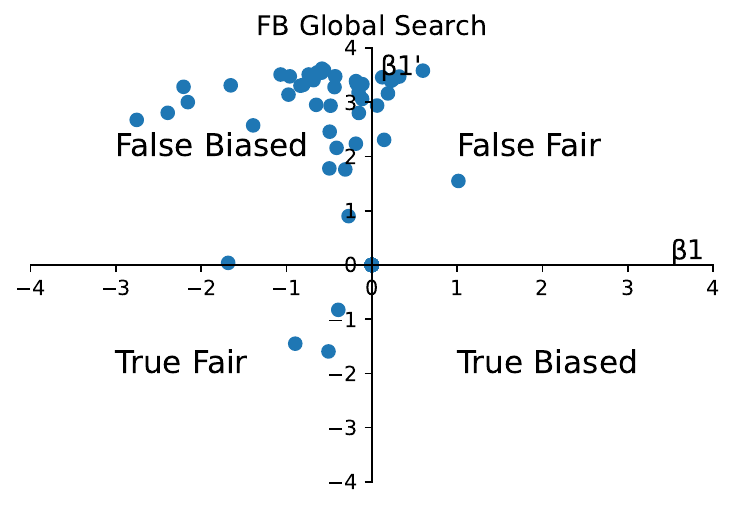}
    \end{minipage}

    \caption{Global Fairness Confusion Perturbation Analysis on COMPAS Dataset}
    \label{compas global analysis}

    \begin{minipage}[t]{0.25\linewidth}
        \centering
        \includegraphics[width=1.0\linewidth]{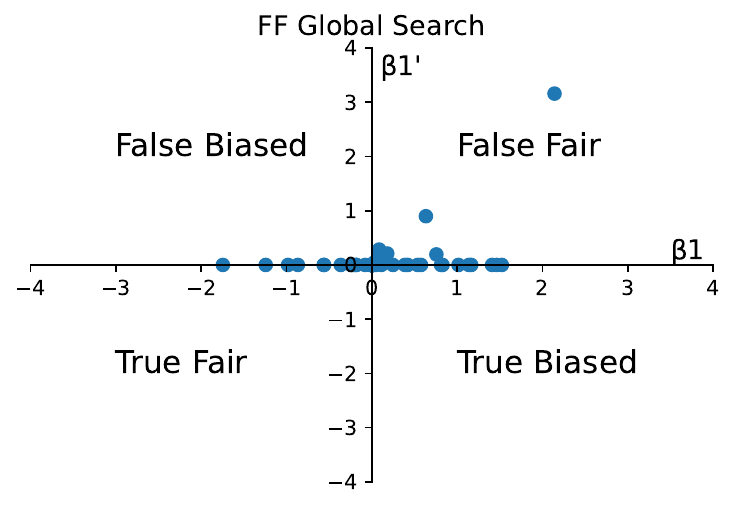}
    \end{minipage}
    \begin{minipage}[t]{0.25\linewidth}
        \centering
        \includegraphics[width=1.0\linewidth]{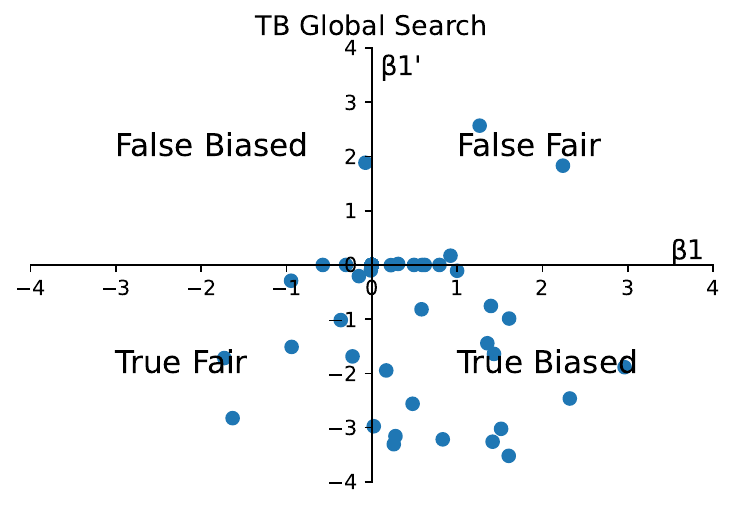}
    \end{minipage}
    \begin{minipage}[t]{0.25\linewidth}
        \centering
        \includegraphics[width=1.0\linewidth]{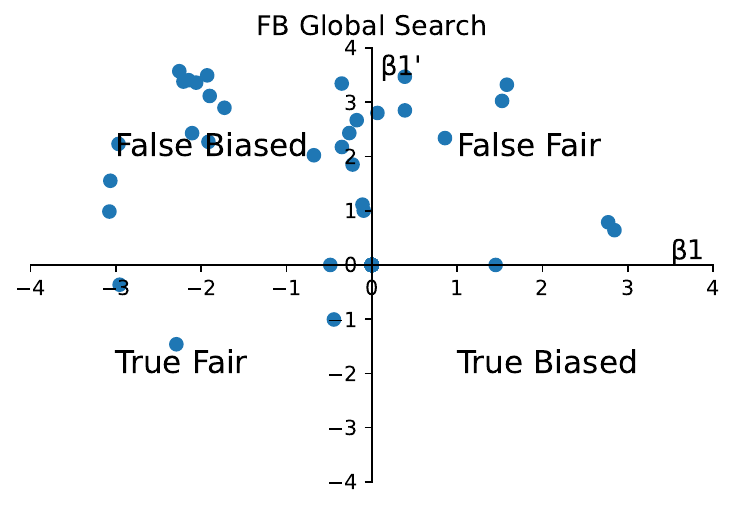}
    \end{minipage}

    \caption{Global Fairness Confusion Perturbation Analysis on German Credit Dataset}
    \label{credit global analysis}
\end{figure}

\begin{figure}[!htb]
\centering
    \begin{minipage}[t]{0.25\linewidth}
        \centering
        \includegraphics[width=1.0\linewidth]{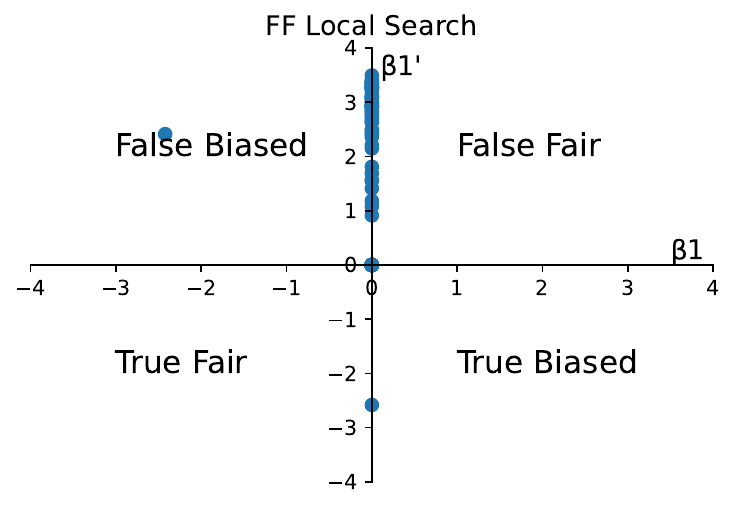}
    \end{minipage}
    \begin{minipage}[t]{0.25\linewidth}
        \centering
        \includegraphics[width=1.0\linewidth]{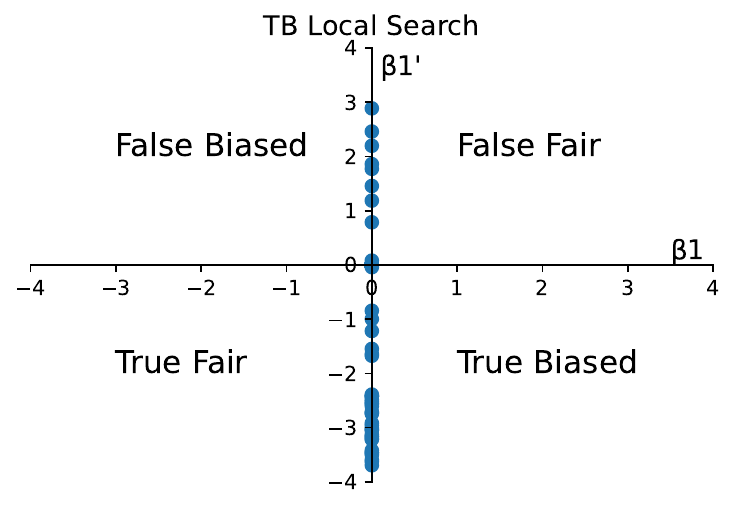}
    \end{minipage}
    \begin{minipage}[t]{0.25\linewidth}
        \centering
        \includegraphics[width=1.0\linewidth]{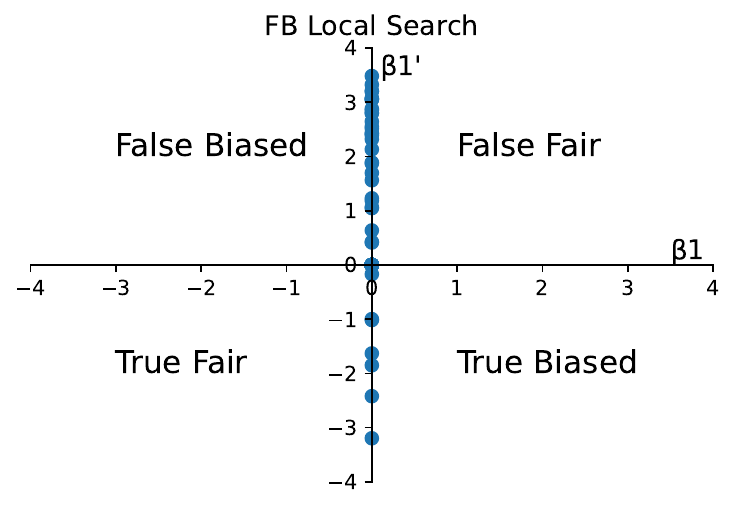}
    \end{minipage}

    \caption{Local Fairness Confusion Perturbation Analysis on Adult Dataset}
    \label{adult local analysis}

\centering
    \begin{minipage}[t]{0.25\linewidth}
        \centering
        \includegraphics[width=1.0\linewidth]{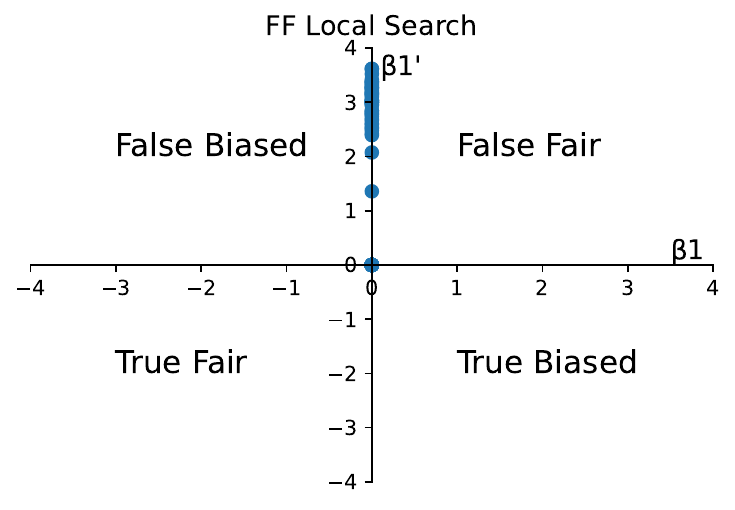}
    \end{minipage}
    \begin{minipage}[t]{0.25\linewidth}
        \centering
        \includegraphics[width=1.0\linewidth]{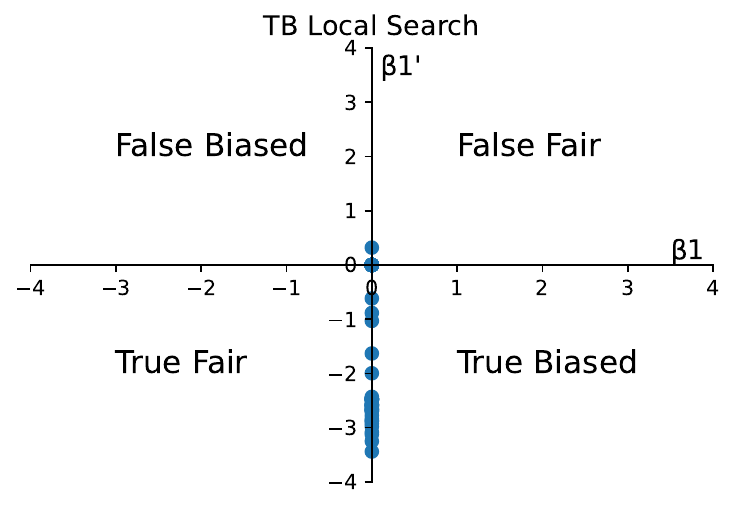}
    \end{minipage}
    \begin{minipage}[t]{0.25\linewidth}
        \centering
        \includegraphics[width=1.0\linewidth]{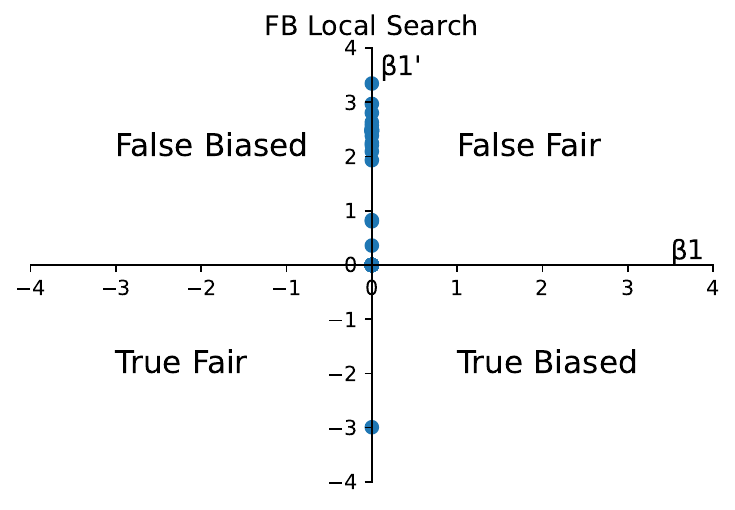}
    \end{minipage}

    \caption{Local Fairness Confusion Perturbation Analysis on COMPAS Dataset}
    \label{compas local analysis}

\centering
    \begin{minipage}[t]{0.25\linewidth}
        \centering
        \includegraphics[width=1.0\linewidth]{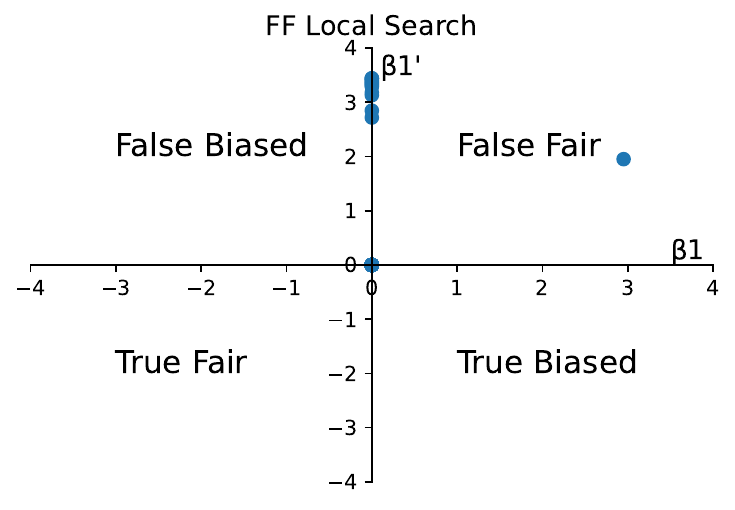}
    \end{minipage}
    \begin{minipage}[t]{0.25\linewidth}
        \centering
        \includegraphics[width=1.0\linewidth]{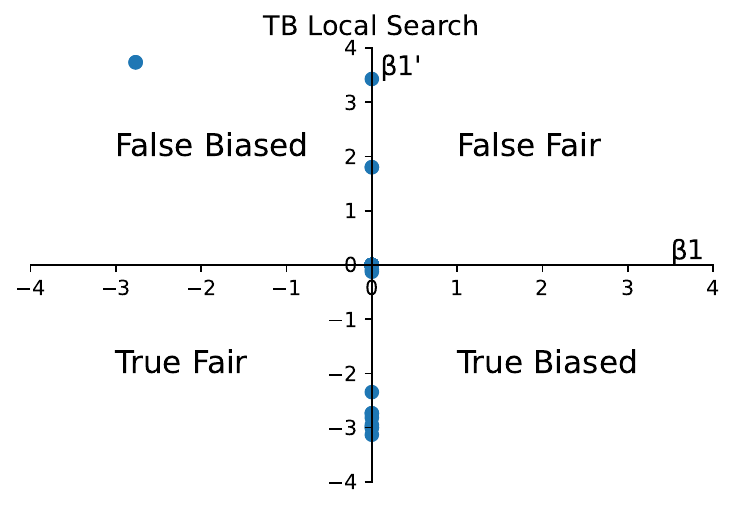}
    \end{minipage}
    \begin{minipage}[t]{0.25\linewidth}
        \centering
        \includegraphics[width=1.0\linewidth]{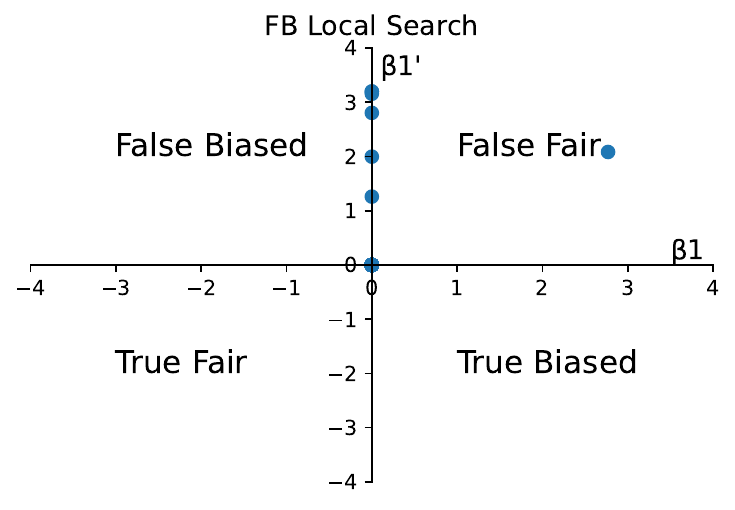}
    \end{minipage}

    \caption{Local Fairness Confusion Perturbation Analysis on German Credit Dataset}
    \label{credit local analysis}
\end{figure}

The statistical results of quantile regression for global fairness confusion perturbations across the Adult, COMPAS, and German Credit datasets are presented in Figures \ref{adult global analysis}, \ref{compas global analysis}, and \ref{credit global analysis}. Similarly, the statistical results of quantile regression for local fairness confusion perturbations across these datasets can be found in Figures \ref{adult local analysis}, \ref{compas local analysis}, and \ref{credit local analysis}. To categorize the coordinate system, we utilize the parameters $\hat{\beta_1}$ and $\hat{\beta_1'}$ derived from the instances and their similar counterparts. This categorization results in four areas: \textit{True Fair} (sign$(\hat{\beta_1})$ = -1, sign$(\hat{\beta_1'})$ = -1), \textit{True Biased} (sign$(\hat{\beta_1})$ = -1, sign$(\hat{\beta_1'})$ = 1), \textit{False Fair} (sign$(\hat{\beta_1})$ = 1, sign$(\hat{\beta_1'})$ = 1), and \textit{False Biased} (sign$(\hat{\beta_1})$ = 1, sign$(\hat{\beta_1'})$ = -1).


As illustrated in Figures \ref{adult global analysis}, \ref{compas global analysis}, and \ref{credit global analysis}, the results indicate that the majority of instances subjected to false fair, true biased, and false biased global perturbations are consistently situated within their respective categories of false fair, true biased, and false biased areas.

In contrast, observing the outcomes of local perturbations in Figures \ref{adult local analysis}, \ref{compas local analysis}, and \ref{credit local analysis}, a noticeable trend emerges where a significant proportion of perturbed instances align along the ordinate axis, with $sign(\hat{\beta_1})=0$. This observation suggests that local perturbations have a relatively minor impact on the instances themselves. However, they notably influence their similar counterparts, thereby effectively identifying false or biased adversarial instances in the proximity of the input instances.

\end{document}